\pdfoutput=1
\documentclass[11pt,table,xcdraw]{article}

\usepackage[final]{acl}

\usepackage{times}
\usepackage{latexsym}

\usepackage[T1]{fontenc}
\usepackage[utf8]{inputenc}

\usepackage{microtype}

\usepackage{inconsolata}

\usepackage{graphicx}

\usepackage{hyperref}       % hyperlinks
\usepackage{url}            % simple URL typesetting
\usepackage{booktabs}       % professional-quality tables
\usepackage{amsfonts}       % blackboard math symbols
\usepackage{nicefrac}       % compact symbols for 1/2, etc.
\usepackage{array}
\usepackage{multirow}
\usepackage{makecell}  % 支持单元格换行
\usepackage{amsmath}
\usepackage{amssymb}
\usepackage{colortbl}
\usepackage[]{xcolor}
\usepackage{wrapfig}
\usepackage{enumitem}
\usepackage{stfloats}
\usepackage{afterpage}
\usepackage{subcaption}

\title{Can Multimodal Large Language Models Understand Spatial Relations?}

\author{
 Jingping Liu\textsuperscript{\rm $\diamondsuit$}\thanks{Corresponding authors.}, 
 Ziyan Liu\textsuperscript{\rm $\diamondsuit$}, 
 Zhedong Cen\textsuperscript{\rm $\diamondsuit$},
 Yan Zhou\textsuperscript{\rm $\diamondsuit$},
 Yinan Zou\textsuperscript{\rm $\diamondsuit$}, \\
 \textbf{Weiyan Zhang\textsuperscript{\rm $\diamondsuit$}\footnotemark[1],
 Haiyun Jiang\textsuperscript{\rm $\spadesuit$},
 Tong Ruan\textsuperscript{\rm $\diamondsuit$}} \\
\textsuperscript{\rm $\diamondsuit$}School of Information Science and Engineering, East China University \\ of Science and Technology, Shanghai, China \\
\textsuperscript{\rm $\spadesuit$}School of Computer Science, Fudan University, Shanghai, China \\
\texttt{\{jingpingliu,weiyanzhang\}@ecust.edu.cn},
\texttt{y30241069@mail.ecust.edu.cn} \\
}

\begin{document}
\maketitle
\begin{abstract}
Spatial relation reasoning is a crucial task for multimodal large language models (MLLMs) to understand the objective world. However, current benchmarks have issues like relying on bounding boxes, ignoring perspective substitutions, or allowing questions to be answered using only the model's prior knowledge without image understanding. To address these issues, we introduce SpatialMQA, a human-annotated spatial relation reasoning benchmark based on COCO2017, which enables MLLMs to focus more on understanding images in the objective world. To ensure data quality, we design a well-tailored annotation procedure, resulting in SpatialMQA consisting of 5,392 samples. Based on this benchmark, a series of closed- and open-source MLLMs are implemented and the results indicate that the current state-of-the-art MLLM achieves only 48.14\% accuracy, far below the human-level accuracy of 98.40\%. Extensive experimental analyses are also conducted, suggesting the future research directions. The benchmark and codes are available at \url{https://github.com/ziyan-xiaoyu/SpatialMQA.git}.
\end{abstract}

% 1. Samples in spatial relation reasoning benchmarks. Q, O,和A in our SpatialMQA 分别表示questions, options, and answers.
% 2. 由于空间限制，我们简写了question and answer。
% 3. A和B的题型是判断题，输入是图片和文字描述，输出是true or false。

\section{Introduction}

Multimodal large language models have become increasingly significant in AI due to their ability to process and integrate data from multiple sources such as text and images. Although MLLMs excel in tasks like image recognition~\cite{guo2023texts} and classification~\cite{wang2023representation}, they still face challenges with more complex tasks, such as multimodal understanding and reasoning~\cite{zheng2023ddcot}, highlighting the need for further exploration and enhancement of their capabilities.

\begin{figure}[t]
\centering
% \vspace{-0.2cm}
\includegraphics[width=0.98\linewidth]{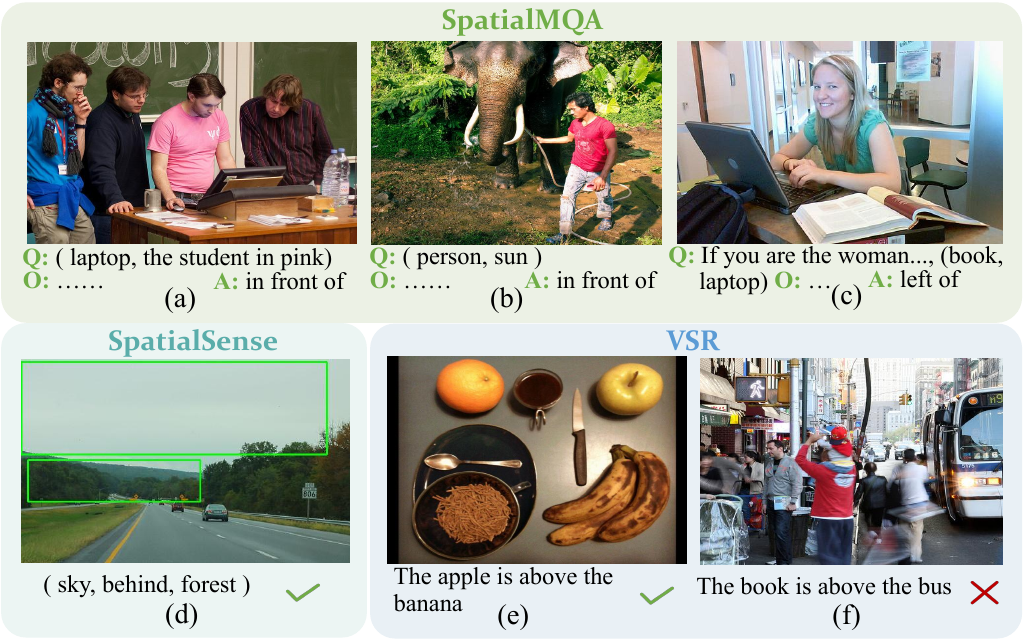}
\caption{Samples from spatial relation reasoning benchmarks. ``Q'', ``O'', and ``A'' in our SpatialMQA denote the question, options, and answer. In SpatialSense~\cite{yang2019spatialsense} and VSR~\cite{liu2023visual}, questions are binary classification, with image and text inputs, and true/false outputs.}
\label{fig:intro}
\vspace{-0.2cm}
\end{figure}

% 1. bboxes
% 2. 坐标系
% 3. 视角带入
% 4. 先验知识

% 但是，尽管MLLMs在一些任务如图像识别和分类表现好，但是在一些复杂任务上如多模态推理上仍表现不佳，导致十分有必要来探寻他们的能力。
% These models excel at comprehending and generating complex information, making them invaluable for applications ranging from NLP and CV to more sophisticated tasks like multi-modal understanding and reasoning. 

% 1. 空间关系的理解是衡量多模态大模型能力的一项重要任务。
% 2. 这个任务是指给定场景中的两个物体，让模型推理出它们的空间关系。
% 3. 例如，如图1a所示，模型需要在给定的图片中推断出云和山的空间关系为“上”。
% 4. 这个任务之所以重要是因为真实世界的空间关系理解是人类在日常生活中所必备的一项基本能力。
% 5. Continuing the previous图片，我们要充分理解图中信息需要知道the objects（a, b, c,d）and their spatial relation (6种关系)。

% 【更改】评估mllm能力的一个关键方面......。例如，如图（1）（a）所示，图像展现了四个人在讲台前研讨的场景。在这个场景图中，模型为了充分理解图像中的信息，需要识别实体(the man in pink, the man in blue, laptop和podium等)以及实体之间的空间关系(laptop on the podium)。其中，一些空间关系会随着观察视角的改变而发生变化，比如，从照片拍摄者的视角来看，the man in blue is left of the man in pink; 从图中the man in pink的视角来看，the man in blue is right of him.

A critical aspect of evaluating MLLMs is their ability to understand spatial relations, which involves inferring the spatial relations between entities in a given scene~\cite{liu2023visual}. For instance, in Figure \ref{fig:intro}(a), given the subject ``laptop'' and the object ``the student in pink'', the model needs to infer that the spatial relation between them is ``in front of''. This task is important because understanding spatial relations in the objective world is a fundamental human ability essential for daily life~\cite{proulx2016relation,hawes2020explains}. For instance, to fully the scene of four students discussing at the podium in Figure \ref{fig:intro}(a), it is necessary to identify the entities (laptop, podium, the student in pink) and their spatial relations (laptop on podium, laptop in front of the student in pink).

\begin{table*}[t]
\centering
\caption{Overview of spatial relation reasoning benchmarks.
``Q. Type'', ``Rel.'', ``Type'', ``Obj. W'', ``Per. sub.'', ``Kn.'', and ``MQA'' stand for ``question type'', ``relations'', ``types of subject or object'', ``objective world'', ``perspective substitution''  ``knowledge'', and ``multiple-choice QA'', respectively. ``Objective world'' indicates whether the benchmark's annotations use the objective world as the reference system. ``Perspective substitution'' means whether questions involve perspective (first- or third-person). 
% ``perspective substitution'' 意味着是否有视角（first and third-person perspective）代入的问题。
``Knowledge'' indicates whether questions in the benchmark can be answered solely with models' prior knowledge, without images.}
% \begin{tabular}{@{}cccccccccc@{}}
\scalebox{0.80}{
% \begin{tabular}{p{5.2cm}p{1.8cm}p{1.6cm}<{\centering}p{0.8cm}<{\centering}p{0.8cm}<{\centering}p{1cm}<{\centering}p{1.1cm}<{\centering}p{1.9cm}<{\centering}p{1cm}<{\centering}p{1.7cm}<{\raggedleft}}
\begin{tabular}{p{5.4cm}p{1.6cm}<{\centering}p{1.1cm}<{\centering}p{1.3cm}<{\centering}p{1.5cm}<{\centering}p{1.4cm}<{\centering}p{1.4cm}<{\centering}p{1.3cm}<{\centering}p{1cm}<{\raggedleft}}
\toprule
\textbf{Benchmark} & \textbf{Q. Type} & \textbf{\# Rel.} & \textbf{\# Type}  & \textbf{w/o bbox} & \textbf{Obj. W} & \textbf{Per. sub.} & \textbf{w/o Kn.} & \textbf{Size} \\ \midrule
% CLEVR$^*$~\cite{johnson2017clevr}     & Various          & 4                    & 96                                            & ×                                & ×    & Outside & -  & - (999,968)                                          \\
SpatialVOC2K~\cite{belz2018spatialvoc2k}     & Cloze          & 17                    & 20                                & ×                 & ×                       & ×         & ×    & 2,026                                             \\
SpatialSense~\cite{yang2019spatialsense}     & T or F                 & 9                     & -                             & ×                 & ×                       & First       & ×            & 17,498                                   \\
Rel3D~\cite{goyal2020rel3d}            & T or F                 & 30                    & 67                                         & ×             & ×            & ×         & -         & 27,336                                       \\
SpatialSense+~\cite{wen2024can}    & T or F                 & 9                     & -         & ×                 & ×        & ×                  & \checkmark      & 7,254                                          \\
SpatialRGPT~\cite{cheng2024spatialrgpt}    & OpenQA                 & 12                     & 88         & ×                 & \checkmark        & ×                  & \checkmark      & 1,406                                          \\
% MME$^*$~\cite{yin2023survey}     & T or F                 & 10                     & 44       &  \checkmark                  & ×        & Outside                  & \checkmark      & 60 (1,457)                                         \\
% EgoThink$^*$~\cite{cheng2023can}     & OpenQA                & -                    & 14         & \checkmark                 & \checkmark        & First                  & \checkmark      & 50 (700)                                         \\
EmbSpatial~\cite{du2024embspatial}     & MQA                & 6                    & 294         & \checkmark                 & ×        & ×                  & ×      & 3640                                         \\
VSR~\cite{liu2023visual}             & T or F                 & 66                    & 32                                   & \checkmark           & Partly                  & First       & ×                   & 10,972                             \\
SpatialVLM~\cite{chen2024spatialvlm}              & OpenQA                 & -                    & -                                   & \checkmark           & Partly                  & ×       & ×                   & 546                             \\
SpatialMQA (ours)      & MQA                   & 6                     & 128                                   & \checkmark          & \checkmark                      & First/Third      & \checkmark             & 5,392                             \\ \bottomrule
\end{tabular}}
\label{tab:1}
\vspace{-0.2cm}
% \end{table}
\end{table*}

Several benchmarks exist for spatial relation reasoning, yet they remain insufficient for fully evaluating MLLMs' ability to understand spatial relations. These benchmarks can be categorized based on whether they use bounding boxes (bboxes) to enclose subjects and objects. First, benchmarks with bbox annotations, such as SpatialVOC2K~\cite{belz2018spatialvoc2k}, Rel3D~\cite{goyal2020rel3d}, and SpatialSense+~\cite{wen2024can}, face two main challenges. On one hand, the subject or object in the question may not be explicitly visible in images, making it impossible to use bboxes~\cite{liu2023visual}. As illustrated in Figure \ref{fig:intro}(b), the subject ``sun'' cannot be framed with a bbox in the question ``Where is the sun located relative to the man?''. On the other hand, some spatial relations in these benchmarks are not grounded in the objective world, leading to a gap between machine and human cognition. For instance, in Figure \ref{fig:intro}(d), the sky is objectively above the forest, but SpatialSense marks it as behind the forest.
% 1. 实际上，空间关系推理的场景是复杂多样的，比如说图像中可能不会显式提及问题中的主体或客体，这导致无法通过bbox来标记。
% 2. 如图1b所示，不可能用bbox来框住the subject “sun” in the question "where is the sun located relative to the man".
% The scenarios for this reasoning are complex and diverse, and certain entities in questions cannot be marked by bboxes. For example, when reasoning about the spatial relation between the sun and the man in Figure \ref{fig:intro}(b), it is impossible to use a bbox to frame the sun, which is not present in the image. 
% 此外，上述数据集的空间关系推理并不是基于客观世界标注的。
% In addition, some spatial relations between entities in the above benchmarks are not grounded in the objective world, leading to a gap between machine and human cognition. For example, in Figure \ref{fig:intro}(c), the sky is objectively above the forest, but SpatialSense marks it as behind the forest. 
Secondly, benchmarks without labeled bboxes, like EmbSpatial~\cite{du2024embspatial}, VSR~\cite{liu2023visual}, and SpatialVLM~\cite{chen2024spatialvlm}, also face two main issues. 
% One major issue is their insufficient focus on perspective substitutions (first- and third-person perspectives). For instance, in VSR, only 6\% of the samples use a first-person perspective.
One major issue is that they often ignore perspective substitution (first- and third-person). Even when included, it is only a small part. For instance, in VSR, only 6\% of the benchmark uses a first-person perspective.
This limits the model's ability to understand spatial relations from different perspectives, which is important for complex, dynamic scenarios like autonomous driving~\cite{gao2024survey}. Another issue is that some questions in these benchmarks can be answered correctly without images, relying only on the model's prior knowledge. As shown in Figure \ref{fig:intro}(f), the question ``the book is above the bus'' can often be answered ``No'' based on commonsense, without needing to analyze the image. This prevents a proper evaluation of MLLMs' image understanding abilities.

Hence, in this paper, we introduce SpatialMQA, a new benchmark in a \textbf{m}ultiple-choice \textbf{q}uestion \& \textbf{a}nswer format, designed to fully evaluate the ability of MLLMs in multimodal \textbf{spatial} relation reasoning. The benchmark includes 5,392 samples based on COCO2017~\cite{lin2014microsoft}, covering 128 subject and object types, without the use of bboxes. To address the limitations of existing benchmarks, we establish clear annotation guidelines for SpatialMQA, incorporating questions that involve perspective substitution based on the objective world as a reference system, while avoiding questions that can be answered solely through the models' prior knowledge without images.
% we clearly define annotation guidelines for SpatialMQA, including standardizing the objective world as the coordinate system and avoiding questions that can be answered solely by models' prior knowledge without images. 
In addition, we design a three-round annotation procedure for quality control. To assess the spatial relation reasoning capabilities of MLLMs, we conduct comprehensive experiments using closed-source models such as GPT-4o~\cite{achiam2023gpt} and Gemini-1.5-flash~\cite{team2023gemini}, as well as open-source models like LLaVA~\cite{liu2024visual} and SpaceLLaVA~\cite{chen2024spatialvlm}. 
% The results indicate that all MLLMs currently struggle with spatial relation reasoning and lack an understanding of the objective world.

In summary, our contributions include:

\begin{itemize}
    \item We introduce a new manually annotated high-quality benchmark for multimodal spatial relation reasoning without bboxes.
    % 1. 我们提出了一个新的人工标注的高质量多模态空间关系推理数据集（SpatialMQA）。该数据集中的问题无法仅凭大模型的先验知识而脱离图像进行回答。

    % 1. SpatialMQA最大的特点是标注的问题涉及基于真实世界为参考系的视角代入和该数据集中的问题无法仅凭大模型的先验知识而脱离图像进行回答。
    \item The main characteristic of SpatialMQA is that the questions involve perspective substitutions using the objective world as a reference. Also, the questions cannot be answered using only the model's prior knowledge without images.
    % SpatialMQA's primary feature is its use of the objective world as a reference system for annotation, involving questions that encompass both the first- and third-person perspectives. The questions in this benchmark cannot be answered using only MLLMs' prior knowledge without the aid of images.

    \item We evaluate both open- and closed-source MLLMs on SpatialMQA, indicating that state-of-the-art (SoTA) methods like GPT-4o and instruction-tuned SpaceLLaVA achieve accuracies of 40.20\% and 48.14\%, respectively, far below the human accuracy of 98.40\%. We further provide detailed analyses and suggest future research directions.

    % We conduct extensive experiments on SpatialMQA and report the performance of both open- and closed-source MLLMs. The accuracy of the state-of-the-art (SoTA) methods, including GPT-4v and the fine-tuned LLaVA, on our benchmark is only 39.80\% and 46.85\%, respectively, which is significantly below the human accuracy level of 97.86\%. We further provide detailed analyses and point out promising directions for future research.
    
    % 1. 目前SoTA方法GPT-4v和微调过的LLaVA在我们数据集上的准确率仅达到了35和47，远远低于人类水平（98的准确率）。
    
\end{itemize}

\section{Problem Formulation}

\label{Problem Formulation}
% 1. 在本文，我们将空间关系识别任务定义为multiple-choice question-answering。
% 2. 给定文本问题Q和图像I，其中Q是关于I中提及到一对目标对象空间关系的问题，空间关系识别（SRR）任务希望模型能从多个选项中选择出正确的选项
% % ，这个选项体现了Q中提及的一对主客体在I中的空间关系。
% 3. 在这里，每个选项是一个空间关系类别，来自于预定义的集合R=（上，下，左，右，前，后），它们的定义如表1所示。
% 4. 比如说，如图1所示，给定Q“XX”和XX.jpg，一个理想的模型会从4个选项（上下左右）中选择出A作为正确答案。
% 5. 需要注意的是，不同问题，选项的个数可能是不一样的。
In this paper, we consider the spatial relation reasoning task as a multiple-choice question-answering problem. Given a text question $Q$ and an image $I$, where $Q$ asks about the spatial relation between two target entities, the task requires the model to select the correct answer from $k \ (k=2, ..., 6)$ options. Each option corresponds to a spatial relation from the pre-defined set $R=\{$\textit{left of}, \textit{right of}, \textit{in front of}, \textit{behind}, \textit{on/above}, \textit{below}$\}$, with their definitions provided in Table \ref{tab:2}. For instance, in Figure \ref{fig:intro}(a), given the question ``Where is the laptop located relative to the student in pink?'', an image, and six options, an ideal model would select ``in front of'' as the correct answer. 
% As an example in Figure \ref{fig:intro}(a), given the question ``Where is the laptop located relative to the student in pink?'', an image, and six options, an ideal model would select ``in front of'' as the correct answer. 

% Note that the number of options may vary for different questions (see Section \ref{SpatialMQA Analysis} for explanations).
% 解释详见4.

% 1. 如图1所示，给定图像（a），问题“Where is the phone located on the laptop?”，以及选项【A，B】，

% \begin{table}[!bt]
\begin{table*}[t]
\centering
\caption{The definition of the spatial coordinate system (SCS) and its six spatial relations. The coordinates for the subject are specified as $(x_s, y_s, z_s)$ and for the object as $(x_o, y_o, z_o)$.}
\scalebox{0.83}{
\begin{tabular}{p{1.6cm}p{16.5cm}}
\toprule
\multicolumn{1}{l}{\textbf{Terms}} & \textbf{Definition} \\ \midrule
\textbf{SCS}         & The spatial coordinate system is established based on the objective world, with gravity pointing downward and the observer as the origin. The X-axis spans the observer's left (negative) to right (positive), the Y-axis from back (negative) to front (positive), and the Z-axis from down (negative) to up (positive). \\ \cmidrule{2-2}
\textbf{left of}        & The subject is to the left of the object when $x_s < x_o$.  \\
\textbf{right of}       & The subject is to the right of the object when $x_s > x_o$.    \\ \cmidrule{2-2}
\textbf{in front of}       & The subject is in front of the object when 1) $y_s \cdot y_o>0$, $|y_s|-|y_o|<0$, or 2) $y_s \cdot y_o<0$, $y_s>0>y_o$.    \\
\textbf{behind}        & The subject is behind the object when 1) $y_s \cdot y_o>0$, $|y_s|-|y_o|>0$, or 2) $y_s \cdot y_o<0$, $y_s<0<y_o$.   \\  \cmidrule{2-2}
% \textbf{Front}       & \begin{tabular}[c]{@{}l@{}}The subject is in front of the object when \\ 1) $y_s*y_o>0$, $|y_s|-|y_o|<0$, or 2) $y_s*y_o<0$, $y_s>0>y_o$.\end{tabular}     \\
% \textbf{Back}        & \begin{tabular}[c]{@{}l@{}}The subject is behind the object when \\ 1) $y_s*y_o>0$, $|y_s|-|y_o|>0$, or 2) $y_s*y_o<0$，$y_s<0<y_o$.\end{tabular}    \\  \cmidrule{2-2}
\textbf{on/above}          & The subject is on/above the object when $z_s > z_o$.        \\
\textbf{below}        & The subject is below the object when $z_s < z_o$.     \\ \bottomrule
\end{tabular}}
\label{tab:2}
\vspace{-0.1cm}
% \end{table}
\end{table*}

\section{SpatialMQA Construction}

\label{SpatialMQA Construction}
% 1. 在本节，我们详细描述了数据集A的构建过程，包括数据来源，标注规范，以及标注流程。
In this section, we detail the construction of the SpatialMQA benchmark, including image source, annotation guidelines, and annotation procedures.

\subsection{Image Source}
\label{Image Source}
% 1. 为了构建真实场景的SpatialMQA数据集，筛选高质量的、实拍的开源图像来源是至关重要的。
% 2. 在本文，我们选择COCO2017作为SpatialMQA的图像来源。
% 3. 之所以选择它，是因为它有以下三个特点：
% 4. 1）庞大的图像数量。COCO2017包含16万多张图像，这对于我们从中挑选出高质量的图像用作空间关系识别是完全足够的。
% 5. 2）丰富的图像类别。COCO2017包含了80种对象类别，覆盖了广泛的日常物体和场景，例如人、车辆、户外物体、动物、运动器材和食物等。
% 6. 3）多对象的图像。COCO2017中的图像通常涉及多个对象，这对于我们选择两个合适的对象用于理解和推理它们的空间关系是更方便的。
% 7. 基于这个数据集，我们随机挑选了\textbf{2万张图像}来标注两个对象及其它们之间的空间关系。
% To construct SpatialMQA with authentic scenes, it is essential to use real-shot open-source images. 
In this study, we choose COCO2017~\cite{lin2014microsoft} as our image source due to its notable advantages: 1) Extensive collection. COCO2017 contains over 160,000 images, providing a broad selection for identifying high-quality images to analyze spatial relations. 2) Diverse types: The dataset encompasses 80 entity types, covering a wide range of entities in the objective world, such as people, animals, cars, and food. 3) Multi-entity scenarios: The images in COCO2017 often involve multiple entities, making it easier to select two appropriate entities to determine their spatial relation. From this dataset, we select 30,000 high-quality images to annotate two entities and their spatial relation.

\subsection{Annotation Guidelines}
\label{Annotation Guidelines}
% 1. 基于上述收集的图片，我们需要对每一张图片进行标注。
% 2. 标注的内容包括对图片的提问，候选项，以及对应的答案。
% 3. 为了让标注者更容易的标注出高质量的样本，我们设计了标注指导，包括问题的类型以及注意事项。
Based on the collected images, we label each one with a question, options, and the correct answer. To assist annotators in creating high-quality samples, we provide annotation guidelines, including question types and important precautions.

% 问题类型。
% 1. 根据观察者的视角，我们将问题分为两大类：图像外和图像内的视角。
% 2. 在第一类中，观察者独立于主体与客体存在于图像外，向其提问图像中两个主客体之间的空间关系。
% 3. 提问的方式主要包括：“主体在客体的左侧（上方）还是右侧（下方）”，“主体在客体的哪一侧”，以及“主体在客体的什么位置”。
% 4. 在第二类中，观察者的视角可以进一步分为两种
% 5. 1）我们将客体视角作为观察者视角，提问图中的某一主体与观察者自身的空间关系。
% 6. 提问的方式主要是“如果你是图中的客体，那么主体在你的左侧（上方）还是右侧（下方）/主体在你的哪一侧/主体在你的什么位置”。
% 7. 2）观察者既不是主体也不是客体，但却是图中的某一生命体。
% 8. 提问的方式主要是“如果你是图中的Role，那么以你的视角看，主体在客体的左侧（上方）还是右侧（下方）/主体在客体的哪一侧/主体在客体的什么位置”。
% 9. 上述问题类型的样例见附录XX。
\textbf{Question types.}
Based on the observer's perspective, we divide the question types into two types: out-of-image and in-image perspectives. In the first type, the observer exists outside the image and we manually pre-define several question templates like ``Where is the subject located relative to the object?''. See Appendix \ref{Question Templates} for more templates. In the second type, the observer’s perspective is within the image and can be further divided into two types. The first type uses the object's perspective as the observer’s perspective (also denoted as the first-person perspective). Question templates for this type address the spatial relation between the subject and the observer (the object), such as ``If you are \textit{[object]} in the image, where is the subject located relative to you?''. The second type considers a living being (third-person perspective) within the image as the observer, distinct from both the subject and the object. It includes question templates like, ``If you are the \textit{[living being]} in the image, from your perspective, where is the subject located relative to the object?''.

% 1. 不清晰、不全
% 2. 模型可以通过常识回答的问题

% 注意事项：
% 1. 注意事项主要目的是指导标注者识别和排除低质量样本。
% 2. 注意事项主要包括三条：
% 3. 第一，图片必须十分清晰，且提问的主体或客体需要在图像中完整的显示出来。
% 4. 第二，提出的问题不能仅依靠先验知识就能回答，比如，披萨在碟子的上方还是下方。
% \textbf{Precautions.}
% This part aims to guide annotators in identifying and excluding low-quality samples. There are two main precautions to consider: First, the subject or object in the image must be clear and complete. Note that some subjects or objects may not be explicitly visible in the image, such as ``sun'' in Figure \ref{fig:intro}(b). Second, the question should not be answerable solely based on prior knowledge without the image, such as ``the book is above the bus'' in Figure \ref{fig:intro}(f).

% This part主要目的是指导标注者识别和排除低质量样本。
% 注意事项主要包括2条：
% 第一，问题不能仅依靠模型的先验知识而不需要图像就能正确回答。比如说，图1中的问题``the book is above the bus''，模型很明显不需要图像就能回答no。
% 第二，图像必须十分清晰，且提问的主体或客体能够很轻易的被识别出来。
\textbf{Precautions.}
This part guides annotators in excluding low-quality samples. There are two main precautions to consider: Firstly, the question cannot be correctly answered based solely on the model's prior knowledge without an image. For instance, the question ``the book is above the bus'' in Figure \ref{fig:intro}(f) can be answered as ``No'' without visual input. Secondly, the image must be clear, with the subject or object of the question being easily identifiable.

\subsection{Annotation Procedure}
\label{Annotation Procedure}
% 1. 为了完成高质量的数据集构建工作，我们组建了一个专业的数据集构建团队，包括标注者三人、检查者三人、以及审核者一人三组成员。
% 2. 针对团队所有成员，我们让他们掌握空间坐标系以及6个空间关系类别的定义，并对他们进行标注规范的培训。
% 3. 以下是数据集构建过程的三个阶段：第一轮标注，第二轮检查，第三轮审核。
% To ensure high-quality benchmark construction, we organize a professional team consisting of three annotators, two checkers, and one reviewer. All team members are trained to understand the definition of the spatial coordinate system, six spatial relations, and comprehensive annotation guidelines. The benchmark construction process involves three stages: first-round annotation, second-round checking, and third-round review.

To create a high-quality benchmark, we organize a professional team of three annotators, two checkers, and one reviewer. All team members are trained to understand the definition of the spatial coordinate system, six spatial relations, and annotation guidelines. The procedure includes first-round annotation, second-round checking, and third-round review.

\begin{table}[t]
\centering
% \caption{Statistics of our constructed SpatialMQA. Min\_L, Max$_$L, Avg$_$L } 
\footnotesize
\caption{Statistics of our SpatialMQA. 
% ``Min\_L'', ``Max\_L'', ``Avg\_L'' represent the minimum, maximum, and average length of questions in the benchmark, respectively. 
The samples in ``first-p''(first-person perspective) and ``third-p''(third-person perspective) are both from ``In-I''(In-image). The latter has fewer samples due to the limited number of images that depict three distinct living entities. ``Out-I'' means ``out-of-image''.}
\scalebox{0.75}{
% \begin{tabular}{p{2.5cm}p{0.8cm}<{\centering}p{0.8cm}<{\centering}p{0.8cm}<{\centering}p{1.0cm}<{\centering}p{1.0cm}<{\centering}p{1.0cm}<{\centering}p{1.0cm}<{\centering}p{1.1cm}<{\centering}}
\begin{tabular}{p{1.4cm}p{0.7cm}<{\raggedleft}p{0.5cm}<{\raggedleft}p{0.6cm}<{\raggedleft}p{0.7cm}<{\raggedleft}p{0.7cm}<{\raggedleft}p{0.62cm}<{\centering}p{0.62cm}<{\centering}p{0.62cm}<{\raggedleft}}
\toprule
                  & \textbf{Train} & \textbf{Dev} & \textbf{Test} & \textbf{Total} & \textbf{Ratio}   & \textbf{Min.L} & \textbf{Max.L} & \textbf{Avg.L} \\ \midrule
SpatialMQA        & 3,780  & 536  & 1,076 & 5,392  & 100.00 & 7           & 34          & 18.84       \\ \cmidrule{1-9}
\rowcolor{gray!15}\multicolumn{9}{c}{\textit{\textbf{Spatial relations}}}        \\ \cmidrule{1-9}
left of           & 1,040  & 148  & 296  & 1,484  & 27.52  & 7           & 34          & 18.39       \\
right of          & 980   & 139  & 279  & 1,398  & 25.93  & 8           & 32          & 18.50       \\
in front of       & 565   & 80   & 161  & 806   & 14.95  & 9           & 33          & 20.04       \\
behind            & 529   & 75   & 151  & 755   & 14.00  & 7           & 34          & 18.61       \\
on/above          & 353   & 50   & 100  & 503   & 9.33   & 8           & 33          & 18.30       \\
below             & 313   & 44   & 89   & 446   & 8.27   & 8           & 32          & 20.29       \\
 \cmidrule{1-9}
\rowcolor{gray!15}\multicolumn{9}{c}{\textit{\textbf{Question types}}}        \\ \cmidrule{1-9}
Out-I      & 1,513  & 217  & 452  & 2,182  & 40.00  & 7           & 33          & 15.81        \\
In-I          & 2,267  & 319  & 624  & 3,210  & 60.00  & 12          & 34          & 20.91       \\
\    \scriptsize {$\#$ \textit{first-p}}   & \scriptsize {2,136}  & \scriptsize {299}  & \scriptsize {590}  & \scriptsize {3,025}  & \scriptsize {94.24}  & \scriptsize {12}         & \scriptsize {34}          & \scriptsize {20.60}      \\
\   \scriptsize {$\#$ \textit{third-p}}   & \scriptsize {131}   & \scriptsize {20}   & \scriptsize {34}   & \scriptsize {185}   & \scriptsize {5.76}   & \scriptsize {18}          & \scriptsize {34}          & \scriptsize {25.91} \\  
 \bottomrule     
\end{tabular}}
\label{tab:statistics}
\vspace{-0.2cm}
% \end{table}
\end{table}
% 第一轮标注。
% 1. 我们邀请了3个大学生，并分别给每个标注者分配1万张图片进行标注。
% 2. 针对每张图片，标注者首先确定是否能提出合适的问题，确保具备空间关系识别所需的多个对象的场景。
% 3. 在符合上述要求的图片中，我们让标注者根据标注规范撰写问题，从预定义集合中选择候选答案，并标记出唯一的正确答案。
\textbf{First-round annotation.}
We invite three college students, assigning 10,000 images to each for annotation. According to the guidelines, they write a reasonable question for each image, select options from a predefined set, and mark the correct answer from the options.

% the annotator first determines whether appropriate questions can be asked to ensure the presence of scenes with multiple entities required for spatial relation recognition. For images that meet these criteria, the annotators write questions according to the annotation guidelines, select options from a predefined set, and mark the correct answer.

% 第二轮检查。
% 1. 标注之后，我们邀请另外两个大学生来检查构建的数据集。
% 2. 一个大学生专门负责检查注意事项1中主客体在图像中是否清晰且完整。
% 3. 另外一个大学生负责检查问题是否能通过先验知识来得到正确答案，在不看图片的前提下。
% 4. 针对检查者认为不合格的样本会被送回去给标注者修改，并附带认为不合格的理由，which serve to re-train the construction team.
% 5. The process repeats until a batch reaches 90 accuracy (i.e., decided to be correct according to the checker).
\textbf{Second-round checking.}
We invite two other college students to simultaneously check the rationality of all samples. Furthermore, each student is assigned an additional task. One student is responsible for checking whether the correct answer to the question can be determined through prior knowledge without images (corresponding to precaution 1). The other student verifies whether the subject or object in the image is clear (corresponding to precaution 2). Samples identified as unqualified by the checkers are returned to annotators with explanations for correction. This process is repeated until a batch achieves 90\% accuracy, as determined by the checkers.

% 第三轮审核
% 1. A verified batch is presented to authors for double-checking.
% 2. Authors conduct random inspections for 20\% samples of a batch, and unqualified annotations are sent back with reasons to the check team to fine-tune their checking criteria, which in turn regularize the construction team. The process also repeats until a batch reaches 95\% accuracy.

% 1. 最后，我们获得了5000条高质量的标注样本。
\textbf{Third-round review.}
A verified batch is given to a main author for double review. The author randomly inspects 20\% of the batch samples. Any unqualified annotations are returned to the check team with explanations, allowing them to refine their criteria, which in turn helps standardize the construction team's work. The cycle continues until the batch achieves 95\% accuracy. Finally, we obtain 5,392 high-quality samples to form SpatialMQA.

% Finally, we compile a benchmark of 5,392 high-quality annotated samples to create SpatialMQA. Each sample includes a question, an image, a variable number of candidate options, and the correct answer.

\section{SpatialMQA Analysis}
\label{SpatialMQA Analysis}

\begin{figure}[t]
\centering
\includegraphics[width=1\linewidth]{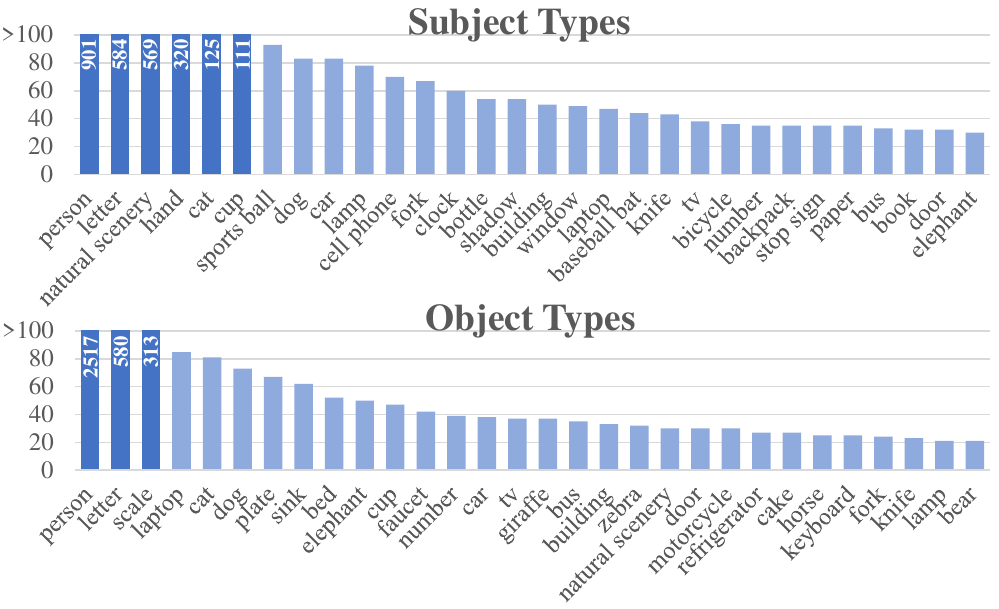}
\caption{Distributions of subject and object types.}
\label{fig:sub}
\vspace{-0.1cm}
% \end{figure}
\end{figure}

\textbf{Benchmark statistics.}
% 1. As reported in Table 1，SpatialMQA包含了5392个样本，其中训练集有3780个，验证有536个，以及测试集有1076个。
% 2. 在这个数据集中，问题的最小长度、最大长度和平均长度分别为A，B和C。
% 2. 基于空间关系的分类，我们观察到左/右关系的样本占比最多，上/下关系的样本占比较低，而前/后关系介于两者之间。
% 3. 此外，X、Y、Z轴的两侧样本数量占比基本均为1:1。
% 4. 基于问题类型的分类，我们观察到以图片外和图片内作为视角的样本比例相当，但进一步，在图片内视角下，客体视角和第三者视角的比例差异较大。
% 5. 此外，在图片内视角下，问题的最小长度是整个数据集问题的最小长度2-3倍。
% 6. 这是因为图片内视角下的问题涉及3个对象，而其余问题一般只涉及2个对象。
As reported in Table \ref{tab:statistics}, SpatialMQA contains 5,392 samples, divided into training, validation, and test sets according to a 7:1:2 ratio. In this benchmark, the questions have a minimum length of 7 words, a maximum length of 34, and an average length of 18.84. 
% Based on the classification of spatial relations, ``left of'' and ``right of'' have the highest proportion of samples, ``on/above'' and ``below'' have a lower proportion, and ``in front of'' and ``behind'' fall in between. 
% In addition, the sample number ratios on both sides of the X, Y, and Z axes are approximately 1:1. 
% Based on the classification of question types, the proportions of out-of-image and in-image perspectives are similar. 
% However, within the in-image perspective, the proportion of object perspective samples significantly differs from third-party perspective samples. 
Notably, the minimum length of questions in the in-image perspective is 2-3 times longer than the minimum length of questions in the entire benchmark, as these questions typically involve three entities, while other questions generally involve only two.

% 问题中主客体类型的多样性
% 1. 为了验证我们数据集问题的多样性，我们对问题中主客体的类型进行了统计。
% 2. 首先，我们利用GPT-4v with in context learning (ICL)来抽取出SpatialMQA中每个问题的主体和客体。
% 3. 其次，我们从整个数据集中随机抽样500个样本，并整理出除COCO2017原始80个类别的常见类别，总共10个。
% 3. 然后，我们利用GPT-4v with in context learning (ICL)来将所有的主客体分类到上述的90个类别。
% 4. 最后，对没有归类到预定义类别中的样本进行人工归类。
% 5. 经统计，我们数据集中的主体类别有113个和一个other类（样本数量小于等于5的主体类别的集合），客体类别有84个和一个other类（样本数量小于等于5的客体类别的集合）。
% 6. 由于主客体类别有重复，主客体类别总计128个。
% 7. 为了更直观的了解主客体类型，我们分别提供了样本数量前50的主客体类别，如图所示。
\textbf{Diversity of subject and object types in questions.}
To verify the diversity of questions in our benchmark, we use GPT-4o with in-context learning (ICL) to extract the subjects and objects in questions and classify them into predefined categories. This process is detailed in Appendix \ref{Statistics of subject and object types}. According to our statistics, there are 113 subject categories and an additional category that includes all subject categories with a sample size of five or fewer, and 84 object categories, along with an additional category that encompasses all object categories with a sample size of five or fewer. Due to the overlap between subject and object types, we have a total of 128 distinct subject and object types. To provide a more intuitive understanding of these types, we present the subject and object types with Top-30 frequency, as shown in Figure \ref{fig:sub}.

% \begin{figure}[htp]
% \centering
% \includegraphics[width=0.98\linewidth]{obj.pdf}
% \caption{Object.}
% \label{fig:obj}
% \end{figure}

% \textcolor{red}{According to} our statistics, there are 113 subject types and one additional type for subjects with five or fewer samples in our benchmark, and 84 object types and one additional type for objects with five or fewer samples. 
% First, we adopt GPT-4v with in-context learning (ICL) to extract the subject and object of each question in SpatialMQA. Second, we randomly select 500 samples from the entire benchmark and manually define common types, in addition to the original 80 types from COCO2017, resulting in a total of 90 types. Third, we employ GPT-4v with ICL to classify every subject and object into these 90 types. Finally, samples that are not classified into predefined types are manually categorized into new types. 

% 选项的组合
% 1. 在SpatialMQA，问题选项的个数是不固定，为了确保选项相对于问题而言是具备合理性的。
% 2. 比如说，“where is the motorcycle located relative to the car?”，的答案显然不应该是“on/above”和“below”，因此这个问题的选项只包括另外四个方向。
% 3. 根据包含维度的数量，我们将选项的个数设置为2（任意一个维度的两个空间关系），4（任意两个维度的四个空间关系）和6。
% 4. 根据我们的统计，问题选项数量为4的样本数为4035个，占比75\%，选项个数为2和6的样本数量分别为637和720，占比为12和13。
\textbf{Option combinations.}
In SpatialMQA, the number of question options varies to ensure they are appropriate for questions. For instance, options like ``on/above'' and ``below'' are not suitable for ``where is the motorcycle located relative to the car?''. Hence, we only include the other four options. Based on the coordinate dimensions, we set the number of options to 2 (two spatial relations in one dimension), 4 (four spatial relations in two dimensions), or 6. According to our statistics, 75\% of samples (4036) have 4 options, while 12\% (637) and 13\% (719) have 2 and 6 options, respectively.

\begin{table*}[t]
\centering
\caption{Model comparison (\%) on our SpatialMQA benchmark. All results are the average of three runs.}
\scalebox{0.8}{
\begin{tabular}{p{2.7cm}p{1.4cm}<{\centering}p{1.1cm}<{\centering}p{1.1cm}<{\centering}p{1.1cm}<{\centering}p{1.1cm}<{\centering}p{1.7cm}<{\centering}p{1.1cm}<{\centering}p{1.1cm}<{\centering}p{1.1cm}<{\centering}p{1.4cm}<{\centering}}
\toprule
\textbf{Model}  & \textbf{Settings} & \textbf{P} & \textbf{R} & \textbf{F1} & \textbf{Acc} & \textbf{Settings} & \textbf{P} & \textbf{R} & \textbf{F1} & \textbf{Acc} \\ \midrule
\rowcolor{gray!15}\multicolumn{11}{c}{\textbf{\textit{Open-source MLLMs}}}  \\  \midrule
BLIP-vqa-base   & -                 & 32.92      & 20.86      & 25.54       & 26.49        & FULL              & 48.12	&31.48	&38.06	&33.64        \\
% BLIP2-opt-2.7B  & -                 & 31.31      & 34.97      & 33.04       & 26.86        & LoRA              & 29.64      & 36.20       & 32.59       & 29.93        \\
BLIP2-opt-2.7B  & -                 & 31.31      & 34.97      & 33.04       & 26.86        & LoRA              & 55.20      & 37.47       & 44.64       & 29.93        \\
InstructBLIP-3B & -                 & 37.42      & 27.47      & 31.69       & 28.53        & LoRA              & 44.22      & 44.80       & 44.51       & 42.38        \\
mPLUG-Owl-7B    & -                 & 34.30	&32.90	&33.58	&26.49       & LoRA              & 36.05	&38.59	&37.28	&31.88        \\
IDEFICS-9B      & -                 & 17.72      & 25.80       & 21.00          & 22.12        & LoRA              & 35.13	&36.41	&35.76	&29.28       \\
LLaVA1.5-7B     & -                 & 30.72	&31.18	&30.95	&29.28      & LoRA              & 46.10	&44.56	&45.32	&46.85       \\
SpaceLLaVA     & -                 & 35.13	&32.58	&33.81	&31.32      & LoRA              & 47.96	&46.18	&47.05	&\textbf{48.14}       \\
\cmidrule{1-11}
\rowcolor{gray!15} \multicolumn{11}{c}{\textbf{\textit{Closed-source MLLMs}}}      \\  \midrule
\multirow{2}{*}{Gemini-1.5-flash}     & 0-shot         & 38.55	&35.47	&36.95	&35.40       & 2-shot     &49.30	&35.11	&41.01	&36.80     \\ 
 & 1-shot         & 51.47	&33.46	&40.55	&36.20        & 3-shot          & 51.52	&35.82	&42.26	&38.00        \\ \cmidrule{2-11}
\multirow{2}{*}{GPT-4o}            & 0-shot         & 48.62	&40.19	&44.01	&\textbf{40.20}        & 2-shot   &   48.70	&38.36	&42.92	&38.40	               \\ 
& 1-shot      &48.04	&39.17	&43.15	&39.00  	      & 3-shot          &  46.76	&36.99	&41.30	&37.80      \\ \cmidrule{1-11}
\rowcolor{gray!15}\multicolumn{11}{c}{\textbf{\textit{Other Methods}}}   \\  \midrule
Random Choose   & -                 & 30.22	&27.97	&29.05	&27.20        & -                 & -          & -          & -           & -            \\
Human           & -                 & 98.56     & 98.40        & 98.48        & \textbf{98.40}      & Text-only         & 23.94      & 24.58     & 24.26       & 24.40\color{red}{\scriptsize (3)}        \\ \bottomrule
% Human           & -                 & 98.27      & 98.05      & 98.16       & \textbf{97.86}        & Text-only         & 21.36      & 21.77      & 21.56       & 22.21\color{red}{\scriptsize (3)}        \\ \bottomrule
\end{tabular}}
\label{tab:main}
\vspace{-0.1cm}
% \end{table}
\end{table*}

% 没有图片情况下的问题的可回答性验证
% Verification of Question Answerable Without Images

\section{Experiments}

In this section, we implement state-of-the-art models on our newly constructed SpatialMQA benchmark, aiming at assessing their performance and identifying the underlying challenges.

\subsection{Baselines}
\label{Methods}
% 1. 在本节，我们主要阐述几类基线方法，包括开源多模态大模型、闭源多模态大模型以及其它。
% In this section, we mainly describe several types of baseline methods, including open-source MLLMs, closed-source MLLMs, and others.

% 1. 为了评估不同大模型在我们新构建的SpatialMQA benchmark上的性能，我们选择了open-source MLLMs, closed-source MLLMs，and 人工做题。
We mainly select three types of methods: open-source MLLMs, closed-source MLLMs, and others.

% 开源多模态大模型
% 1. 我们采用了BLIP、BLIP2、InstructBLIP、LLaVA、mPLUG-Owl、IDEFICS for comparison。
% 2. 基于上述模型，我们采用了两种设置：直接推理和微调。
% 3. 在第一种设置下，模型的输入是一张图像、一个问题、以及多个选项，其输出是一个正确选项。
% 4. 需要注意的是，除了BLIP，其余模型都提供了instruction作为输入的一部分。
% 4. 在微调设置下，由于BLIP的参数量较少和对比的其余开源多模态大模型参数量较多，我们对BLIP采用全量微调的方式，对其余开源多模态大模型采用参数高效微调的方式with LoRA。
% 5. 指令数据是通过训练数据的输入和输出改造得到的，instructions与上一种设置保持一致，具体见附录。
\textbf{Open-source MLLMs.}
We use BLIP~\cite{li2022blip}, BLIP2~\cite{li2023blip}, InstructBLIP~\cite{dai2024instructblip}, mPLUG-Owl~\cite{ye2023mplug}, IDEFICS~\cite{laurenccon2023introducing}, LLaVA~\cite{liu2024visual}, and SpaceLLaVA~\cite{chen2024spatialvlm} for comparison. Two settings are designed: direct inference and instruction tuning. In the first setting, models directly produce the answer given an image, a question, and multiple options. Note that all models except BLIP receive a task prompt, as described in Appendix \ref{Details of Open-source MLLMs}.
% Note that all models except BLIP provide instructions as part of the input. 
In the second setting, we use different tuning strategies: full parameter updates for BLIP and parameter-efficient tuning (LoRA~\cite{hu2021lora}) for the other MLLMs.
Instruction data is generated by transforming the input and output from the training data, and the task prompt remains consistent with the previous setting.

% 闭源多模态大模型
% 1. 我们采用了目前最强大的Gemini和GPT-4v for experiments。
% 2. 针对这两个模型，我们同样采用了两种设置：zero-shot reasoning and few-shot reasoning。
% 3. 在第一种设置下，我们只输入图像、问题和选项，并利用instruction让大模型输出答案，其中instruction详见附录。
% 4. 在第二种设置下，我们分别设定1-，2-，3-shot ICL，with the same instruction above。
% 5. 其中，ICL的样本是从训练集中随机挑选的。
\textbf{Closed-source MLLMs.}
We randomly select 500 samples in SpatialMQA and adopt Gemini-1.5-flash and GPT-4o, two of the most powerful models, for our experiments. For both models, we employ two settings: zero-shot reasoning and few-shot reasoning. In the first setting, we input images, questions, and options, and use a task prompt to guide the MLLM to output answers. Detailed prompts are provided in Appendix \ref{Details of Closed-source MLLMs}. In the second setting, we use 1-shot, 2-shot, and 3-shot ICL with the same instructions. The ICL examples are randomly selected from the training set and fixed for all samples in the test set. 
% \textcolor{red}{Since these two MLLMs} are limited to a single image input, we manually convert the images in the ICL examples into text descriptions in advance~\cite{liang2024scemqa}.
% 1. 由于这两个MLLMs限制了只能输入一张图像，因此，我们会事先通过人工的方式将ICL中图像转化为文本描述【文献】。

% 其它
% 1. 我们还设计了两类方法：随机选择和人工作答。
% 2. 在第一种方法中，我们采用python的random函数来为每个题目随机中选项中选择一个答案,
% 2. 在第二种方法中，我们首先随机选取了500个样本，并邀请3名大学生（与标注志愿者不是同一批人）来分别对这些样本的问题进行回答。
% 3. 然后，我们采用major voting方式来综合不同回答者的结果。
% 4. 当三个大学生的答案都彼此不相同时，我们认为这道题回答错误。
\textbf{Other methods.}
We further design two methods: random selection and manual answering. In the first method, we use a random function to select an answer from the options for each question. In the second method, we randomly select 500 samples from SpatialMQA and invite three college students (different from the annotation team in Section \ref{Annotation Procedure}) to answer the questions. The final answer is determined by majority voting, and if the three students provide different answers, the question is considered incorrect.

\subsection{Settings and Metrics}
The hyperparameter settings for the open-source MLLMs are detailed in Appendix \ref{Hyperparameter}. These models are executed on a workstation with two NVIDIA A100-PCIE-40GB GPUs. In the experiments, we report four metrics: precision (P), recall (R), F1, and accuracy (Acc).

% Q1, Q2, and Q3 represent "perspective Out-of-image", "first-person perspective In-image", and "third-person perspective In-image". Ax, Ay, and Az represent answers involving left of and right of on the X-axis, in front of and behind on the Y-axis, and on/above and below on the Z-axis
\begin{table*}[t]
\centering
\caption{Results (Acc \%) grouped by question types and answer types. Q1, Q2, and Q3 represent the question from the ``Out-of-image'' perspective, ``first-person'' , and ``third-person'' perspectives in images. Ax, Ay, and Az represent answers involving ``left of'' and ``right of'' on the X-axis, ``in front of'' and ``behind'' on the Y-axis, and ``on/above'' and ``below'' on the Z-axis. $^\dag$ and $^\ddag$ denote the best few-shot settings in the Main Results, specifically 3-shot and 1-shot, respectively. }
\scalebox{0.8}{
\begin{tabular}{p{3cm}p{2.5cm}<{\centering}p{1.8cm}<{\centering}p{1.8cm}<{\centering}p{1.8cm}<{\centering}p{1.8cm}<{\centering}p{1.8cm}<{\centering}p{1.8cm}<{\centering}}
\toprule
\textbf{Model}                       & \textbf{Settings}  & \textbf{Q1} & \textbf{Q2} & \textbf{Q3}& \textbf{Ax} & \textbf{Ay} & \textbf{Az} \\ \midrule
\rowcolor{gray!15} \multicolumn{8}{c}{\textbf{\textit{Open-source MLLMs}}}  \\ \cmidrule{1-8}
BLIP-vqa-base       & Full       &40.93	&36.10 	&52.94      & 39.65 	&25.64 	&28.57 				     \\
BLIP2-opt-2.7B      & LoRA        & 32.30        & 28.47	&23.53      & 11.65       &49.04      & 53.97                       \\
InstructBLIP-3B     & LoRA      & 44.47       & 40.68       & 44.12         & 36.17       & 48.72       & 50.79                     \\
mPLUG-Owl-7B    & LoRA      &37.83	&28.14	&17.65            & 17.74 	&46.47 	&50.79 				  \\
IDEFICS-9B     & LoRA       &33.41	&26.95	&14.71           & 15.13 	&45.51 		&45.50 			  \\
LLaVA1.5-7B     & LoRA     &53.14	&40.99	&64.71            & 55.71 	&29.64 	&48.13 				   \\
SpaceLLaVA     & LoRA     &54.87	&42.37	&58.82            & 56.00 	&51.85 	&31.41 				   \\
                                     \cmidrule{1-8}
% \multicolumn{2}{c}{Average}                             & 18.84          & 36.16          & 36.97         & 30.05       & 25.45       & 17.65       \\ \cmidrule{1-8}
\rowcolor{gray!15} \multicolumn{8}{c}{\textbf{\textit{Closed-source MLLMs}}}      \\ \cmidrule{1-8}
\multirow{2}{*}{Gemini-1.5-flash}        
% & text-only            &     32.08 		&29.00 		&20.62 		&29.09	&26.83	&26.47 \\
                                     & 0-shot      &42.73	&26.83	&50.00       & 39.17 	&26.25 	&41.00 				                   \\ 
                                     & Few-shot$^\dag$     &48.18	&26.83	&52.94         & 49.58 	&21.88 	&36.00 				                  \\ \cmidrule{2-8}
\multirow{2}{*}{GPT-4o}            
% & text-only            &   26.25 		&16.00 		&36.88 		&21.82	&28.86	&55.88                 \\
                                     & 0-shot     &44.09	&33.74	&61.76        &   37.08 	&47.50 	&36.00 				       \\
                                     & Few-shot$^\ddag$      &45.00	&32.52	&47.06         &   38.75 	&38.75 		&40.00 			                     \\ \cmidrule{1-8}
\rowcolor{gray!15}\multicolumn{8}{c}{\textbf{\textit{Other Methods}}}      \\ \cmidrule{1-8}
% \multicolumn{2}{c}{Random Choose} 
Random Choose & -      & 30.00	& 24.80	&26.47    & 25.42 &27.50 			&31.00 		                            \\  \cmidrule{2-8}
% \multirow{2}{*}{Human}               & Text-only                      & 21.24    & 22.37       & 23.53                & 21.74                              & 24.36  & 18.52     \\ 
%                                      &-      & 98.51                & 97.31                & 100.00    & 97.89     & 97.16     & 99.08                      \\
\multirow{2}{*}{Human}               & Text-only           & 25.91                & 23.17               & 23.53           & 23.75    & 25.00       & 25.00                        \\ 
                                     &-      & 98.51                &  98.24            & 100.00    & 98.61     & 97.79    & 98.68                     \\
\bottomrule
\end{tabular}
}   
\vspace{-0.2cm}
\label{tab:Group analysis}
% \end{table}
\end{table*}
% \section{Experiments}
% % 实验
% % 1. 参数设置（多次实验，平均结果，实验机器）（附录）
% % 2. 指标（附录）
% % 3. 主实验
% % 4. 分析实验
% % 5. case
% In this section, we conduct extensive experiments to evaluate the capabilities of MLLMs on our constructed SpatialMQA benchmark. In addition, we provide detailed insights and analyze the error types of MLLMs to identify future research directions.

% \subsection{Settings and Metrics}
% In the experiments, we report four metrics: precision (P), recall (R), F1, and accuracy (Acc). In addition, the hyperparameters in open-source MLLMs are detailed in Table \ref{tab:Hyperparameter}. These models are executed on a workstation equipped with an Intel(R) Xeon(R) Gold 6348@2.60GHz, 512G memory, and two NVIDIA A100-PCIE-40GB GPUs.

% Model comparison (\%) on our SpatialMQA. All results are the average of three results.
% \begin{table}[t]

\subsection{Main Results}
\label{Main Results}
% 1. 我们在SpatialMQA上运行了第5章提及的所有方法。
% 2. 实验结果如表4所示。
% 3. 从结果中，我们可以得出以下结论：
% 4. 1）所有的模型在SpatialMQA上表现的性能都不理想，与人类水平（准确率97.86）相比还有很大的提升空间。
% 5. 性能最好的模型SpaceLLaVA with LoRA，这个模型是在LLaVA基础上通过大量spatial VQA data微调得到的，也只有48.14\%的准确率，其中LLaVA的visual instruction tuning也涉及integration of coordinate data with bboxes and corresponding captions。这说明了我们构建的SpatialMQA对于MLLM是具有挑战性的。
% 6. 2）在Open-source MLLMs中，微调的模型比不微调的模型在空间关系推理的性能上表现更好。
% 7. 特别地，微调后的LLaVA相比于不微调的LLaVA在准确率上提升了17.57.
% 8. 此外，模型参数多并不意味着空间关系推理能力强。
% 9. 比如说，IDEFICS-9B是所有对比的开源MLLMS中参数量最大的，但其在SpatialMQA上的效果是最差的in both two settings.
% 10. 可能的原因是该模型预训练的数据涉及空间关系推理任务的较少。
% 11. 3）在closed-source LLMs中，GPT-4V with 0-shot 表现是最好的，但随着ICL的样本数量增加，它的准确率会下降。
% 12. 相反，Gemini是随着ICL样本数量的增加，准确率上升。
% 13. 可能的原因是Gemini相比于GPT-4V有较好的基于ICL例子描述的图像场景勾勒能力。
% 14. 4）在Other methods中，人类在只通过问题和选项，没有图像，来作答时，在准确率上有22.21（随机从测试集中挑选了500个样本），低于随机选择。
% 15. 这说明了我们的数据集必须得依赖图像来作答。
% 16. 也就是说，我们的数据集鲜有只依靠先验知识就能回答的问题。
% 17. 通过人工标注，发现500个样本中仅有3个只依靠先验知识就能回答的问题。
We perform all baseline methods on our SpatialMQA benchmark. The experimental results are presented in Table \ref{tab:main}. From the table, we notice that:
1) All MLLMs perform poorly on SpatialMQA, with significant room for improvement compared to the human accuracy of 98.40\%. The best-performing model, SpaceLLaVA with LoRA, achieves only 48.14\% accuracy, despite being fine-tuned on LLaVA with a large amount of spatial VQA samples. Notably, LLaVA’s visual instruction tuning also involves incorporating coordinate data with bboxes and corresponding captions. This indicates that our SpatialMQA benchmark presents a significant challenge for MLLMs. 2) Among open-source MLLMs, instruction-tuned models excel in spatial relation reasoning compared to those without instruction tuning. For instance, the instruction-tuned SpaceLLaVA shows a 16.82\% accuracy improvement over its non-instruction-tuned version. Among closed-source LLMs, GPT-4o performs best with zero-shot learning, but its accuracy decreases as the number of ICL samples increases. In contrast, Gemini's accuracy improves with more ICL samples. The reasons for these opposing results are explained in ``Impact of different ICL examples'' of Section \ref{Detailed Analysis}. 
% 1. 这两个现象规律相反的原因见第5.4章的“Impact of different ICL examples”
3) In other methods, when humans answer questions without images, the accuracy is 24.40\% (based on a random selection of 500 samples from the test set), which is comparable to random selection and significantly lower than the accuracy achieved with images. This indicates that our benchmark heavily relies on images to answer questions. In other words, our benchmark rarely includes questions that can be answered solely with prior knowledge. Furthermore, manual annotation reveals that only 3 out of 500 samples could be answered using prior knowledge alone.

\subsection{Detailed Analysis}
\label{Detailed Analysis}
% 问题种类以及答案种类的分组分析。
% 1. 我们在第3.2章提及了问题类型包含``Out-of-image''（记为Q1）和``In-image''(``first-person view''记为Q2和``third-person view''记为Q3)。
% 2. 此外，我们也对答案类型进行了分类，包括Ax，Ay，Az，分别表示答案在左右、前后、和上下的。
% 3. 上述分组的实验如表5所示。
% 4. 从表中，我们可以得到两个结论：
% 5. 相对于人类在不同维度的空间关系推理能力都是相当的而言，所有的模型在不同的分组上有着明显的性能差异。
% 6. 比如说，人类在Ax，Ay，Az上的结果都在98左右，但是SpaceLLaVA with LoRA在这些分组上的最大差距达到了24.59.
% 7. 这说明了均衡的提升模型在不同空间关系上的推理能力也是至关重要的。
% 5. 1）相比于人类而言，所有的模型在分组的不同维度上体现出了明显的能力差异。
% 6. 相对于人类而言，不同维度的空间关系推理能力都是相当的（如97.89 in Ax，99.08 in Ay，97.16 in Az）。
% 7. 因此，提升模型的空间关系推理能力要均衡。
% 8. 2）在所有的维度上，微调后的模型往往比没微调的模型表现更好，除了BLIP在Ax和BLIP2在Az上有细微的反常情况。

\textbf{Group analysis of question types and answer types.}
As mentioned in Section \ref{Annotation Guidelines}, question types include ``Out-of-image'' (denoted as Q1) and ``In-image'' (further divided into ``first-person perspective'' (Q2) and ``third-person perspective'' (Q3)). In addition, we classify the answer types as Ax, Ay, and Az, representing answers involving \textit{left of} and \textit{right of} on the X-axis, \textit{in front of} and \textit{behind} on the Y-axis, and \textit{on/above} and \textit{below} on the Z-axis, respectively. The results are listed in Table \ref{tab:Group analysis}. 
From the table, we observe that 
% 1) \textcolor{red}{Compared to humans}, all models show significant differences in abilities across different dimensions. For humans, spatial relation reasoning abilities are fairly consistent across dimensions (e.g., 98.61\% in Ax, 97.79\% in Ay, 98.68\% in Az). Therefore, improving model performance should be balanced across these dimensions. 
human reasoning abilities in spatial relations are generally consistent across different groups, but all models display significant performance discrepancies within these groups. For instance, human scores for Ax, Ay, and Az are consistently around 98\%, while SpaceLLaVA with LoRA exhibits a maximum performance gap of 24.59\% in these groups. This suggests that it is essential to improve the model's reasoning abilities in various spatial relations in a balanced manner.
% 2) In all dimensions, instruction-tuned models generally perform better than non-instruction-tuned models, with slight anomalies observed in BLIP on Ax and BLIP2 on Az.

% \begin{table}[t]
\begin{table*}[t]
    \centering
    \begin{minipage}{.5\textwidth}
        \centering
        \footnotesize
        \caption{Results (Acc \%) for different ICL.}
        \begin{tabular}{p{2.1cm}p{1.2cm}<{\centering}p{1.2cm}<{\centering}p{1.2cm}<{\centering}}
            \toprule
            \textbf{Model}          & \multicolumn{1}{c}{\textbf{Settings}} & \textbf{Alig.}         & \textbf{Misalig.} \\ \midrule
            \multirow{3}{*}{Gemini-1.5-flash}& 1-shot   & 36.42             & 35.16        \\
                                     & 2-shot   & 37.14             & 35.94             \\
             & 3-shot   & 38.28             & 37.09         \\  \cmidrule(l){2-4}
            \multirow{3}{*}{GPT-4o}  & 1-shot   &39.09        & 37.75  \\
                                   & 2-shot   & 39.43  & 36.96        \\
             & 3-shot   &  39.88            & 35.87   \\  \bottomrule
            \end{tabular}
            \label{tab:ICL}
    \end{minipage}%
    \begin{minipage}{.5\textwidth}
        \centering
        \footnotesize
        \renewcommand{\arraystretch}{1.07}
        \caption{Impact (Acc \%) of images and option counts.}
        % ``I'', ``Q'', and ``O'' represent the image, question, and options in the task input, respectively.}
        \begin{tabular}{p{2.1cm}p{1.2cm}p{1.2cm}<{\centering}p{1.2cm}<{\centering}}
            \toprule
            \multicolumn{2}{c}{}         & \textbf{All} & \textbf{Part} \\ \midrule
            \multirow{1}{*}{Random}     
                                          & - & 17.20   & 27.20            \\ \cmidrule(l){2-4} 
            \multirow{2}{*}{Gemini-1.5-flash}       & Q+O   & 23.20 	&27.60             \\
                                          & I+Q+O & 29.60   &35.40             \\ \cmidrule(l){2-4} 
            \multirow{2}{*}{GPT-4o} & Q+O   & 26.40 	&27.80             \\
                                          & I+Q+O & 33.80   &40.20             \\ \bottomrule
            \end{tabular}
            \label{tab:ablation}
    \end{minipage}
    \vspace{-0.2cm}
% \end{table}
\end{table*}

% 不同ICL样本的结果。
% \begin{wraptable}{r}{0.5\textwidth}
% % \begin{table}[!bt]
% \centering
% \vspace{-0.4cm}
% \caption{\textcolor{red}{Results} (Acc \%) for different ICL.}
% \scalebox{0.8}{
% \begin{tabular}{p{2.0cm}p{1.2cm}<{\centering}p{1.2cm}<{\centering}p{1.2cm}<{\centering}}
% \toprule
% \textbf{Model}          & \multicolumn{1}{c}{\textbf{Settings}} & \textbf{Alig.}         & \textbf{Misalig.} \\ \midrule
% \multirow{3}{*}{Gemini-Pro-V}& 1-shot   & 34.62             & 33.76        \\
%                          & 2-shot   & 36.32             & 35.04             \\
%  & 3-shot   & 36.75             & 35.90         \\  \cmidrule(l){2-4}
% \multirow{3}{*}{GPT-4V}  & 1-shot   &37.18        & 35.47  \\
%                        & 2-shot   & 38.03  & 33.76        \\
%  & 3-shot   &  38.89            & 32.05   \\  \bottomrule
% \end{tabular}
% \label{tab:ICL}
% }
% % \end{table}
% \end{wraptable}

% 不同ICL样本的影响。
% 1. 在前面闭源大模型的实验中，我们引入了ICL。
% 2. 为了探究不同ICL对模型的影响，我们基于问题类型将ICL分为了两类。
% 3. 一是ICL样例与输入的问题类型是对齐的；二是ICL样例与输入的问题类型是不对齐的。
% 4. 为此，我们随机从500个样本中分别选择问题类型是Q1，Q2和Q3的100个样本（若某类不足以100样本时，我们全部取出。）
% 5. 实验结果表6所示
% 6. 从表中，我们观察到：
% 7. 1）模型with对齐的ICL样例比模型with不对齐的ICL样例在性能上表现更好。
% 8. 比如说，GPT-4V with aligned 3-shot ICL examples比GPT-4V with midaligned 3-shot ICL examples在准确率上提升了6.84.
% 9. 2）在对齐的情况下，GPT-4V的空间关系推理能力会随着ICL样例数量的增加而提升，这是与表4中的结论相反。

\textbf{Impact of different ICL examples.}
We introduce ICL samples for closed-source MLLMs in experiments. To explore the impact of different ICL examples, we divide them into two categories: aligned with the input question type and misaligned. For evaluation, we randomly selected 100 samples for question types Q1, Q2, and Q3 respectively (if a certain category has fewer than 100 samples, we use all available samples). The results are listed in Table \ref{tab:ICL}. From the results, we notice that models with aligned ICL examples outperform those with misaligned ICL examples. For instance, GPT-4o with aligned 3-shot ICL examples improves accuracy by 4.01\% over misaligned ones. 
Notably, the decrease in GPT-4o's spatial relation reasoning ability, mentioned in Section \ref{Main Results}, may be due to the misalignment of examples with the input question type. In contrast, Gemini's performance improves with more ICL examples in the misaligned setting. This could indicate that Gemini effectively utilizes a wider range of examples to enhance generalization and extract relevant features despite the misalignment.

\textbf{Impact of images and option counts.}
% \textcolor{red}{To explore} the impact of images (I) and option (O) counts, 
We conduct analysis experiments by either removing images (I) in the input or using a fixed count of six options (O). The results are listed in Table \ref{tab:ablation}. From the results, we draw the conclusions: 1) MLLMs with Q+O, when tested with varying options, perform similarly to random selection and significantly underperform MLLMs with I+Q+O. This indicates that our benchmark heavily relies on image inputs and cannot depend solely on the model’s prior knowledge. 2) MLLMs with Q+O still perform significantly better than random selection (17.20\%) when given a fixed set of six options. This is because some of the options in this set contradict common sense, allowing the model to exclude them, even without image inputs. This observation is why we remove options that contradict commonsense from our benchmark.
% MLLMs with Q+O perform similarly to random selection，且远远低于MLLMs with I+Q+O, indicating that our benchmark relies on images and cannot rely solely on the model's prior knowledge. 2) MLLMs with Q+O have significantly higher performance compared to random selection (17.20\%) 在固定6个选项的设置下。这是因为固定的6个选项中有不符合常识的选项，导致尽管只有问题和选项，大模型也不会选择那些不符合常识的选项。这也是我们在数据集中会删除不符合常识选项的原因。

% MLLMs perform better with varied options than with a fixed number. However, the significantly higher performance compared to random selection (17.20\%) suggests the presence of unreasonable options, leading us to choose the partial options setting.

% \begin{wrapfigure}{r}{0.5\textwidth}
% \centering
% \vspace{-0.3cm}
% \includegraphics[width=0.95\linewidth]{1.pdf}
% \caption{Distribution of error types.}
% \vspace{-0.2cm}
% \label{fig:distribution}
% \end{wrapfigure}

\begin{figure}[t]
\centering
\includegraphics[width=\columnwidth]{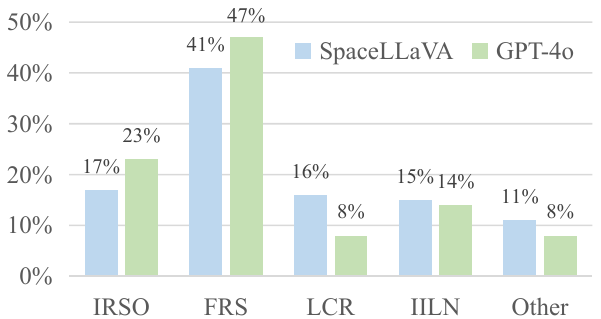}
\caption{Distribution of error types.}
\label{fig:distribution}
\vspace{-0.2cm} % 调整这个值来控制空白的大小
\end{figure}

\subsection{Error Types}
\begin{figure*}[!t]
\centering
\includegraphics[width=0.98\linewidth]{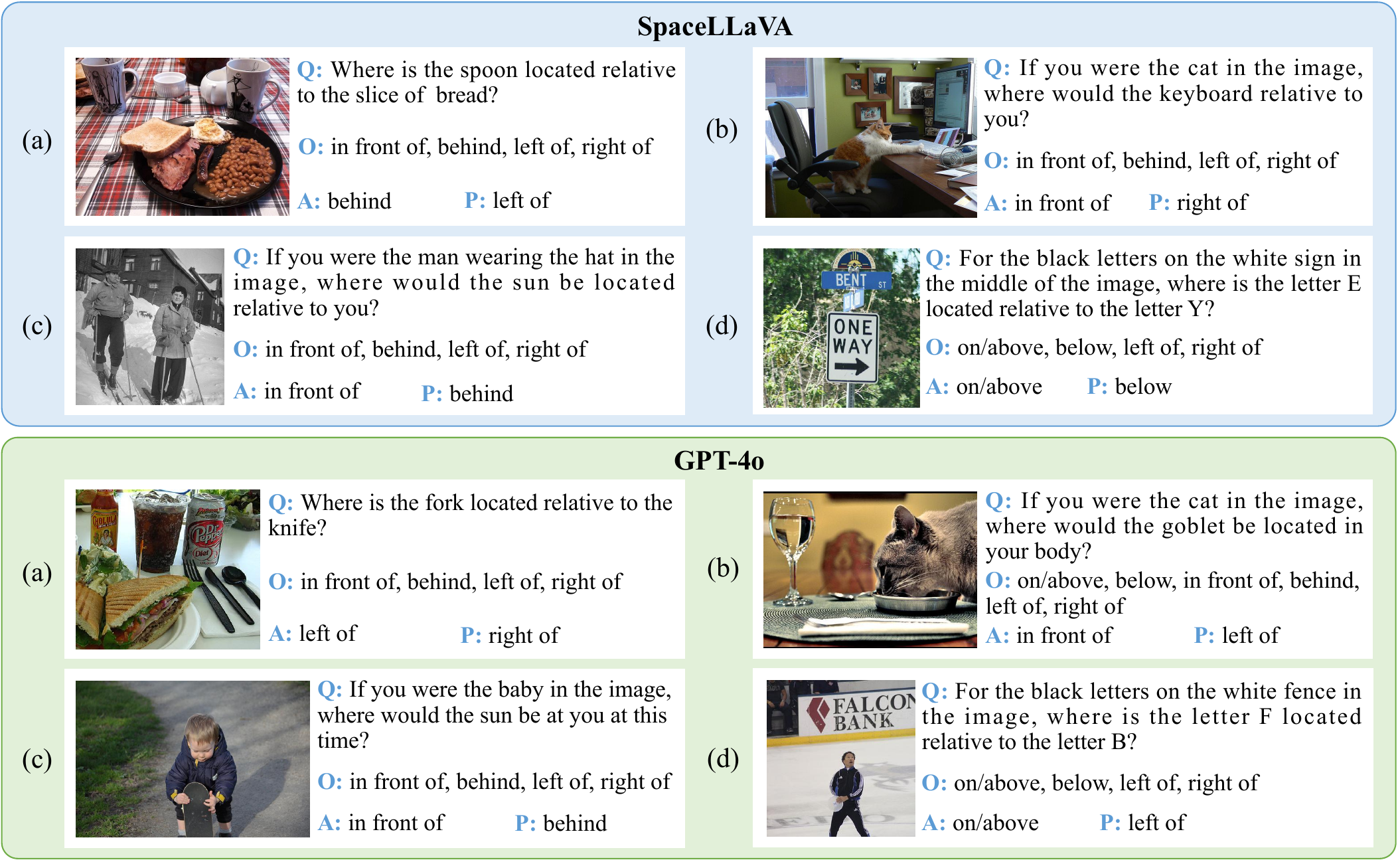}
\caption{Error examples. (a), (b), (c), and (d) describe examples of the IRSO, FRS, LCR, and IILN types, respectively. ``A'' and ``P'' represent the ground truth answer and predicted answer.}
\label{fig:case}
\vspace{-0.2cm}
% \end{figure}
\end{figure*}

% 1. 为了进一步确定MLLMs在空间关系推理上的未来研究方向，我们提供了错误类型的分析。
% 2. 我们分别挑选了LLaVA和GPT-4V的200个错误样本，其中这两个模型是在我们SpatialMQA上表现最好的开源大模型和闭源大模型。
% 3. 经过人工对归类，MLLMs的错误类型包括主客体的错误识别（记为IrSO），角色代入失败，缺乏常识推理能力，字母和数据的和others。
% After manual classification, the error types of MLLMs include incorrect recognition of subjects and objects (IRso), failure in role substitution (Frs), lack of commonsense reasoning (Lcr), incorrect identification of spatial relationships for letters and numbers (IIln), and other errors.
% 4. 具体的错误分布如图3所示。
% 5. 从图中，我们观察到Frs这类错误的数量最多，IRso、Lcr和IIln错误的数量相当。
% 6. 为了更直观的理解这些错误类型，我们提供了样例，如图4所示。
To guide future research in spatial relation reasoning for MLLMs, we analyze 200 error samples produced by SpaceLLaVA and GPT-4o on SpatialMQA. After manual classification, error types are divided into four categories and other errors: (a) incorrect recognition of subjects and objects (IRSO), (b) failure in perspective substitution (FRS), (c) lack of commonsense reasoning ability (LCR), and (d) incorrect identification of spatial relations for letters and numbers (IILN). The error distribution is shown in Figure \ref{fig:distribution}. We observe that FRS errors are the most frequent, with IRSO, LCR, and IILN errors being comparable. To illustrate these error types more intuitively, we provide examples, as shown in Figure \ref{fig:case}.

% 错误类型的分布情况。Examples of error types.

% (a)~(d)代表的错误类型详见5.5节
% \begin{figure}[!bt]

\section{Related Work}
\textbf{Spatial relation reasoning.}
% 1. 图像中主客体之间的空间关系推理是理解真实世界的一项重要任务。
% 2. 目前研究该任务的数据集主要分为两类：with and without bboxes.
% 3. 依赖bboxes的数据集可以进一步分为人工合成和现实场景。
% 4. 在人工合成的数据集中，主要包括CLEVR和Rel3D。
% 5. 但是该类数据集不能准确反映客观世界的真实场景，因此基于真实场景的数据集陆续被提出来，主要包括SpatialVOC2K、SpatialSense、SpatialSense+、SpatialRGPT-Bench、NLVR2、COCO、GQA等。
% 6. 但是，这些数据集仍局限于bbox的标注，存在两个问题：
% 7. 一是一些复杂的空间关系推理无法通过bbox来标注，二是导致模型对图像语义理解的门槛降低，因为bbox在一定程度上可以辅助模型较好的完成空间关系推理任务。
% 8. 因此，VSR数据集without bbox被构建出来。但是该数据集在标注的过程中存在标注标准不统一的情况。
% 9. 此外，该数据集主要是站在第三人称视角进行标注的，涉及第一人称视角的提问不足6\%。
% 10. 相关的数据集还包括MME和EgoThink，其中前者仅考虑第三人称视角，后者是一个多模态的数据集，涉及空间关系推理的数据不足100条。
% 【修改补充】
% 10. 相关的数据集还包括EmbSpatial-Bench、MME、SpatialVLM和EgoThink，其中前三者仅考虑第三人称视角，且并未全部基于真实的客观世界进行标注；SpatialVLM的benchmark数据量较少，且至今仍未开源；EgoThink仅涉及第一人称视角，作为一个多模态数据集涉及空间关系推理的数据不足100条。
Identifying spatial relations between subjects and objects in images is crucial for understanding the world. Benchmarks for this task fall into two main types: those with bboxes and those without. 
The former are sourced from either synthetic or real-world scenes.
% benchmarks that rely on bboxes can be further classified into artificially synthesized and real scenes. 
CLEVR~\cite{johnson2017clevr} and Rel3D~\cite{goyal2020rel3d} are typical examples of synthesized benchmarks, but they do not accurately reflect real-world scenes. Hence, several real-scene benchmarks have been proposed, including SpatialVOC2K~\cite{belz2018spatialvoc2k}, SpatialSense~\cite{yang2019spatialsense}, SpatialSense+~\cite{wen2024can}, 
SpatialRGPT-Bench~\cite{cheng2024spatialrgpt},
NLVR2~\cite{suhr2017corpus}, COCO~\cite{lin2014microsoft}, and GQA~\cite{hudson2019gqa}. 
However, they still use bboxes, which cause two problems: first, some complex spatial relations can’t be fully captured with bboxes~\cite{liu2023visual}; second, bboxes make it easier for models to solve tasks without fully understanding the image~\cite{wen2024can}.
% However, these benchmarks still rely on bbox annotations, which present two issues: first, complex spatial relations cannot always be annotated with bboxes~\cite{liu2023visual}; second, bboxes lower the threshold for the model to understand image semantics, as they assist in completing spatial relation reasoning tasks~\cite{wen2024can}. 
% To overcome these limitations, \textcolor{red}{the VSR benchmark}~\cite{liu2023visual}, which does not use bboxes, is constructed. However, this benchmark faces inconsistent annotation standards and some questions can be answered using prior knowledge without images. Other relevant benchmarks include MME~\cite{Fu2023MMEAC}, which only considers the third-person perspective, and EgoThink~\cite{cheng2023can}, a multi-task benchmark with fewer than 100 samples involving spatial relation reasoning.
% 1. 沿着后者路线的典型数据集包含EmbSpatial-Bench、MME、SpatialVLM、EgoThink和VSR。
% 2. 前面三个仅考虑了out-of-image perspectives，且并没有着重考虑基于真实的客观世界进行标注。
% 3. 需要注意的是，SpatialVLM的测试数据集较小，且至今仍未开源。
% 4. EgoThink仅考虑第一人称视角，且涉及的空间关系推理数据集不足100条。
% 5. VSR虽然考虑了不同视角，但是数据量不足整个数据集大小的6\%，且它的一些questions can be answered using prior knowledge without images. 
Typical benchmarks for the latter include EmbSpatial-Bench~\cite{du2024embspatial}, MME~\cite{Fu2023MMEAC}, SpatialVLM~\cite{chen2024spatialvlm}, EgoThink~\cite{cheng2023can}, and VSR~\cite{liu2023visual}. The first three focus only on out-of-image perspectives and do not always annotate samples based on the objective world. Note that SpatialVLM’s test set is small and not yet open-sourced. EgoThink is limited to the first-person perspective, with fewer than 100 samples in its spatial reasoning benchmark. While VSR considers different perspectives, only 6\% of its data covers them, and some questions can be answered using prior knowledge without images.

% 1. bboxes
% 2. 坐标系
% 3. 视角带入
% 4. 先验知识

% this benchmark faces the problem of inconsistent annotation standards. Additionally, it is primarily annotated from a third-person perspective, with questions involving the first-person perspective accounting for less than 6\%. 

\textbf{Multimodal large language models.} 
% 1. 随着多模态大模型的发展，陆续有很多研究学者将多模态大模型应用到多模态下游任务中，并不断刷新这些任务的性能上限。
% 2. 目前的MLLMs可以分为闭源与开源模型两类。
% 3. 典型的闭源模型主要包括GPT-4V和Gemini。
% 4. 将它们适配到多模态任务中主要有两种方式：包括ICL和Chain-of-Thought。
% 5. 典型的开源大模型主要包括BLIP2、LLaVA、和mPLUG等。
% 6. 由于参数量较少的开源大模型在指令跟随能力上略有不足，将它们适配到下游任务中主要依靠微调的形式，主要有两种：全量参数的更新和非常少量参数的更新，如LoRA和P-tuning v2.
% 8. 尽管目前大模型取得了promising的进展，但仍在我们构建的SpatialMQA数据集上表现欠佳。
With the development of MLLMs, many researchers have applied these models to multimodal downstream tasks. MLLMs can be divided into two categories: closed- and open-source models. Typical closed-source MLLMs include GPT-4o and Gemini. Common methods to adapt these models for multimodal tasks mainly include ICL~\cite{shukor2023beyond,liu2023mmhqa} and Chain-of-Thought (CoT)~\cite{zhang2024cocot,wang2024t}. Typical open-source MLLMs include BLIP2~\cite{li2023blip}, LLaVA~\cite{liu2024visual}, and SpaceLLaVA~\cite{chen2024spatialvlm}. Due to their relatively limited instruction-following capabilities, open-source MLLMs often require instruction tuning for downstream tasks. This tuning can involve full parameter updates or minimal parameter updates, such as LoRA~\cite{hu2021lora} and P-tuning v2~\cite{liu2021p}. Despite the promising progress of current MLLMs, they still perform poorly on our constructed SpatialMQA benchmark.

\section{Conclusion}
\label{Conclusion and Limitations}
% 【limitation补充】
% 1. 数据集的局限性：SpatialMQA被提出的目标在于，评估MLLMs在空间关系推理任务上的表现。作为基准数据集，SpatialMQA的size是足够的。因为数据集内容的特殊性（例如，包含大量的视角代入问题），目前现有的自动化标注工具[1,2,3]均不适用于它，因此目前数据集的构建过程仍依赖于人工。未来，我们将在后续工作中努力探索、开发适用于扩展此种类型数据的自动化标注工具，并根据SpatialMQA的标准，构建足够用于MLLMs训练的数据集。
% 2. 实验评估的可拓展：我们的数据集排除了先验，通过消融图像的实验，能够验证视觉信息带来的增益，进而验证了MLLMs的空间关系推理能力；但我们未涉及模型训练中存在的非故意的数据泄露，未来我们会尝试使用新的评估指标，评估模型依赖预训练知识的情况，从而进一步检验并讨论MLLMs推理空间关系的能力。

% 1. 作为基准数据集够了，但是训练大模型不够，因为都是人工精标的。
% 2. 关系是基本的，可以扩更多。

We introduce SpatialMQA, a manually annotated multimodal spatial relation reasoning benchmark based on COCO2017. To address the weaknesses of existing benchmarks, SpatialMQA is constructed without bboxes, involving perspective substitutions based on the objective world and excluding questions that can be answered solely by model's prior knowledge without images. We implement a series of closed- and open-source MLLMs and conducted extensive experimental analyses. The results indicate that SpatialMQA is a challenging benchmark worth further exploration. 

% However, the benchmark has a limitation: a long tail in the distribution of subject and object types, which mirrors real-world scenarios and can be addressed through machine learning methods~\cite{cao2019learning} or data sampling strategies~\cite{chawla2002smote}.

\section*{Limitations}
% While SpatialMQA presents a valuable benchmark for current MLLMs, it has two key limitations:
% 1. 我们提出SpatialMQA的初衷是为了评估MLLMs在空间关系推理任务中的表现。
% 2. 为了保证benchmark的高质量，我们采用了人工精心标注的方式。
% 3. 这种方式尽管能保证测试集有足够的规模，但是无法提供海量的高质量数据来充分训练MLLMs。
% 4. 尽管目前有些文献[1,2,3]提及的自动化标注工具，但这些都难以适配到我们的benchmark中，因为我们的benchmark主要涉及复杂样例 based on the 真实客观世界 from multiple perspectives。
% 1. SpatialMQA目前仅涵盖六种基本的空间关系（on/above, below, in front of, behind, left of, right of），尚未包含更复杂的关系。原因在于，实验结果表明，这六种基本关系已对当前的MLLMs构成了显著挑战。我们希望MLLMs能够首先掌握这些基本关系的推理能力，为处理更复杂的任务奠定坚实基础。

While SpatialMQA offers a valuable benchmark for evaluating current MLLMs, it has two main limitations: 1) SpatialMQA is created to assess the performance of MLLMs in spatial relation reasoning. To ensure high data quality, we design a manual annotation process, which guarantees a well-constructed and reliable test set. However, this method limits the scale of the training set, making it insufficient for fully fine-tuning MLLMs. Although several automatic annotation tools for spatial relation reasoning are mentioned in \cite{chen2024spatialvlm,cheng2024spatialrgpt,cai2024spatialbot}, they are unsuitable for SpatialMQA due to its complex real-world samples from multiple perspectives. 2) SpatialMQA currently covers six basic spatial relations (left of, right of, in front of, behind, on/above, and below), and does not include more complex relations. We focus on these six because experimental results show they already pose significant challenges to current MLLMs. Mastering these fundamental relations is essential before tackling more complex spatial reasoning tasks.

% 1. 尽管我们新构建的SpatialMQA对于目前的MLLMs来说是具有挑战性，但是它仍然有两个局限：1）SpatialMQA被提出的目标在于，评估MLLMs在空间关系推理任务上的表现。它作为基准数据集，数据量是足够的。其训练集的规模，也能够支撑一些开源模型的微调，但是想要充分地训练大参数的MLLMs还是不够的，因为目前数据集的规模受到人工成本的限制。由于现有的自动化标注工具[1,2,3]均不适用于SpatialMQA（因为内容的特殊性，比如，包含大量的视角代入问题），所以人工成本在该数据集构建过程中依旧需要着重考虑。2）SpatialMQA目前考虑的空间关系数量有限。但这六种空间关系（on/above, below, in front of, behind, left of, right of），是现实场景中最常见和最基本的空间关系。我们希望一个模型能够先具备准确推理这六种关系的能力，以具备在实际情况中的应用基础。如果这些基本推理问题不能得到有效解决，那么处理更细粒度或更复杂的空间关系可能会更具挑战性。正如我们的实验结果所显示的，SpatialMQA对于目前的MLLMs仍具有较大的挑战性，我们希望SpatialMQA能够激发对提升MLLMs空间关系能力的进一步研究。未来，我们也会随着模型性能的提升，在后续的研究中，覆盖更为复杂的空间关系。

\section*{Ethical Statement}
% 1. 我们的SpatialMQA数据集是基于COCO2017【】构建的，which is licensed under a Creative Commons Attribution 4.0 License.
% 2. 因此，我们将SpatialMQA的版权设置为CC-BY 4.0 license.
% 3. 此外，我们仔细检查了我们数据集中的问题，并没有涉及到一些有害问题，比如男女歧视，种族歧视，黄色问题等。
Our SpatialMQA benchmark is built upon COCO2017~\cite{lin2014microsoft}, which is licensed under the Creative Commons Attribution 4.0 License. This license allows us to distribute and re-annotate the dataset, as long as the original work is properly cited. Hence, we release SpatialMQA under the CC-BY 4.0 license. Additionally, we have carefully reviewed the benchmark to ensure it contains no harmful content, such as gender bias, racial discrimination, or inappropriate material.
% 1. 这个许可允许我们散播和重新标注，只要我们引用了它们的工作。

\section*{Acknowledgments}
This paper was supported by the National Natural Science Foundation of China (No. 62306112), Shanghai Sailing Program (No. 23YF1409400), and Shanghai Pilot Program for Basic Research (No. 22TQ1400100-20).

\bibliography{main}

\clearpage

\appendix

\section{Question Templates}
\label{Question Templates}
% 1. 在第3.2章中，我们提出了三种问题类型。
% 2. 针对这些类型，我们人工定义了一些问题模版，如表8所示。

In Section \ref{Annotation Guidelines}, we design three types of questions. For each type, we manually define several question templates, as listed in Table \ref{tab:templates}. Q1, Q2, and Q3 indicate that the sample's question type is ``Out-of-image'', the first-, and third-person perspective of ``In-image'', respectively.

\vspace{-0.1cm}
\begin{table}[!h]
\centering
\caption{Question Template.}
\vspace{-0.2cm}
\scalebox{0.85}{
\begin{tabular}{p{0.5cm}p{7.5cm}}
\toprule
\textbf{}                        & \textbf{Question Template}                                                              \\ \midrule
\multirow{3}{*}{\textbf{Q1}} & Is \underline{××} located to the left or right of \underline{××}?             \\
                                 & Which side of \underline{××} is \underline{××} located on?         \\
                                 & Where is \underline{××} located relative to \underline{××}?            \\ \cmidrule(l){2-2} 
\multirow{3}{*}{\textbf{Q2}}          & If you are \underline{××} in the image, is \underline{××} located to your left or right?                              \\
                                 & If you are \underline{××} in the image, which side of \underline{××} is \underline{××} located on?    \\
                                 & If you are \underline{××} in the image, where is \underline{××} located relative to you?                      \\ \cmidrule(l){2-2} 
\multirow{3}{*}{\textbf{Q3}}          & If you are \underline{××} in the image, from your perspective, is \underline{××} located to the left or right of \underline{××}?  \\
                                 & If you are \underline{××} in the image, from your perspective, which side of \underline{××} is \underline{××} located on?          \\
                                 & If you are \underline{××} in the image, from your perspective, where is \underline{××} located relative to \underline{××}? \\ \bottomrule
\end{tabular}}
\label{tab:templates}
\end{table}
\vspace{-0.2cm}

\section{Statistics of Subject and Object Types}
\label{Statistics of subject and object types}
In Section \ref{SpatialMQA Analysis}, we use GPT-4o with ICL to extract the subjects and objects in questions and classify them into predefined categories. The process is as follows. First, we adopt GPT-4o with ICL to extract the subject and object of each question in SpatialMQA. Second, we randomly select 500 samples from the entire benchmark and manually define common types, in addition to the original 80 types from COCO2017, resulting in a total of 90 types. Third, we employ GPT-4o with ICL to classify every subject and object into these 90 types. Finally, samples that are not classified into predefined types are manually categorized into new types.

\section{Details of Open-source MLLMs}
\label{Details of Open-source MLLMs}
% 1. 在第5章中，我们考虑了open-source MLLMs作为基线模型。
% 2. 这些模型的task prompts和instruction data format如表9和10所示。
In Section \ref{Methods}, we consider open-source MLLMs as baseline models. The task prompts and instruction data format of these models are presented in Tables \ref{tab:promptO} and \ref{tab:instructionF}.

\begin{table}[!t]
% [!bt]
\centering
\caption{Task prompts for open-source MLLMs.}
\vspace{-0.1cm}
\scalebox{0.85}{
\begin{tabular}{p{1.8cm}p{6.2cm}}
\toprule
\textbf{Models}          & \textbf{Task prompt}           \\ \midrule
BLIP & Input: Image: <image>, Question: \{question\}, Options: \{options\}. \textbackslash n Output:            \\       \cmidrule(l){2-2}          
BLIP2, InstructBLIP, IDEFICS, mPLUG-Owl, LLaVA, SpaceLLaVA         & You are currently a senior expert in spatial relation reasoning. \textbackslash n Given an Image, a Question, and Options, your task is to answer the correct spatial relation. Note that you only need to choose one option from all options without explaining any reason. \textbackslash n Input: Image: <image>, Question: \{question\}, Options: \{options\}. \textbackslash n Output:          \\ \bottomrule
\end{tabular}}
\label{tab:promptO}
\end{table}

\begin{table}[!t]
% [!bt]
\centering
\caption{Instruction data format for open-source MLLMs.}
\vspace{-0.1cm}
\scalebox{0.85}{
\begin{tabular}{p{1.8cm}p{6.2cm}}
\toprule
\textbf{Models}          & \textbf{Instruction}           \\ \midrule
BLIP & Input: Image: <image>, Question: \{question\} Options: \{options\}. \textbackslash n Output: \{answer\} \\       \cmidrule(l){2-2}          
BLIP2, InstructBLIP, IDEFICS, LLaVA, SpaceLLaVA         & You are currently a senior expert in spatial relation reasoning. \textbackslash n Given an Image, a Question, and Options, your task is to answer the correct spatial relation. Note that you only need to choose one option from all options without explaining any reason. \textbackslash n Input: Image: <image>, Question: \{question\}, Options: \{options\}. \textbackslash n Output: \{answer\} \\      \cmidrule(l){2-2}          
mPLUG-Owl         & The following is a conversation between a curious human and an AI assistant. \textbackslash n Human: <image> \textbackslash n Human: You are currently a senior expert in spatial relation reasoning. 
 \textbackslash n Given an Image, a Question, and Options, your task is to answer the correct spatial relation. Note that you only need to choose one option from all options without explaining any reason. \textbackslash n Input: Image: <image>, Question: \{question\}, Options: \{options\}. \textbackslash n Output: \textbackslash n A: \{answer\}   \\ \bottomrule
\end{tabular}
}
\label{tab:instructionF}
\end{table}

\section{Details of Closed-source MLLMs}
\label{Details of Closed-source MLLMs}
In Section \ref{Methods}, we consider closed-source MLLMs as baseline models. The task prompts of these models are listed in Table \ref{tab:promptC}.

\begin{table}[!t]
% [!bt]
\centering
\caption{Task prompts for closed-source MLLMs.}
\scalebox{0.85}{
\begin{tabular}{p{1.5cm}p{6.5cm}}
% \begin{tabular}{p{2cm}p{15.5cm}}
\toprule
\textbf{}          & \textbf{Task prompt}           \\ \midrule
Zero-shot & You are currently a senior expert in spatial relation reasoning. \textbackslash n Given an Image, a Question, and Options, your task is to answer the correct spatial relation. Note that you only need to choose one option from all options without explaining any reason. \textbackslash n Input: Image: <image>, Question: \{question\}, Options: \{options\}. \textbackslash n Output: \\       \cmidrule(l){2-2}          
Few-shot         & You are currently a senior expert in spatial relation reasoning. \textbackslash n Given an Image, a Question, and Options, your task is to answer the correct spatial relation. Note that you only need to choose one option from all options without explaining any reason. \textbackslash n Given the following 3 examples to learn the spatial relation reasoning task: \textbackslash n Example1: Input: Image: <image> \textbackslash n Question: For the clock in the image, does the hour hand point above or below the 9 scales?, Options: on/above; below. \textbackslash n Output: above. \textbackslash n Example2: ... \textbackslash n Example3: ... \textbackslash n Input: Image: <image> \textbackslash n Question: \{question\}, Options: \{options\}. \textbackslash n Output: \\      \cmidrule(l){2-2}          
Text-only         & You are currently a senior expert in spatial relation reasoning. \textbackslash n Given an Image, a Question, and Options, your task is to answer the correct spatial relation. Note that you only need to choose one option from all options without explaining any reason. \textbackslash n Input: Question: \{question\}, Options: \{options\}. \textbackslash n Output:  \\ \bottomrule
\end{tabular}
}
\label{tab:promptC}
% \vspace{-0.1cm}
\end{table}

\section{Hyperparameter Settings}
\label{Hyperparameter}
Details of the hyperparameter settings for open-source MLLMs are presented in Table \ref{tab:Hyperparameter}.

\begin{table}[!t]
\centering
\vspace{-0.2cm}
\caption{Hyperparameter settings for open-source MLLMs. ``Ep'', ``BS'', ``ES'', ``LR'', ``Opt'', ``LR. S'', ``PAW8'', ``ExpLR'' and ``LD'' stand for ``Epochs'', ``Batch Size'', ``Early Stop'', ``Learning Rate'', ``Optimizer'', ``LR Schedule'', ``Paged\_Adamw\_8bit'', ``ExponentialLR'', and ``Linear Decay'', respectively.}
\scalebox{0.8}{
% \begin{tabular}{p{2.3cm}p{2.3cm}<{\centering}p{2.3cm}<{\centering}p{2.3cm}<{\centering}p{2.0cm}<{\centering}p{2.3cm}<{\centering}p{3cm}<{\centering}}
\begin{tabular}{p{2.2cm}p{0.4cm}<{\centering}p{0.4cm}<{\centering}p{0.4cm}<{\centering}p{0.7cm}<{\centering}p{1.4cm}<{\centering}p{1.2cm}<{\centering}}
\toprule
\textbf{Model} & \textbf{Ep} & \textbf{BS} & \textbf{ES} & \textbf{LR} & \textbf{Opt} & \textbf{LR. S} \\  \midrule
BLIP & 30 & 8 & 5 & 6e-7 & AdamW & ExpLR \\ 
BLIP2 & 30 & 8 & 5 & 4e-5 & AdamW & ExpLR \\ 
InstructBLIP & 30 & 8 & 5 & 4e-5 & AdamW & ExpLR \\ 
mPLUG-Owl & 10 & 8 & - & 5e-5 & AdamW & LD \\ 
IDEFICS & 10 & 8 & - & 2e-4 & PAW8 & LD \\ 
LLaVA & 10 & 8 & - & 2e-4 & AdamW & Cosine \\ 
SpaceLLaVA & 10 & 8 & - & 2e-4 & AdamW & Cosine \\ \bottomrule     
\end{tabular}
}
\label{tab:Hyperparameter}
% \end{table}
\end{table}

% % \begin{table}[!h]
% \begin{table*}[t]
% \centering
% \caption{Hyperparameter settings for open-source MLLMs.}
% \scalebox{0.8}{
% \begin{tabular}{p{2.3cm}p{2.3cm}<{\centering}p{2.3cm}<{\centering}p{2.3cm}<{\centering}p{2.0cm}<{\centering}p{2.3cm}<{\centering}p{3cm}<{\centering}}
% \toprule
% \textbf{Hyper-para.} & \textbf{BLIP} & \textbf{BLIP2} & \textbf{InstructBLIP} & \textbf{LLaVA} & \textbf{mPLUG-Owl} & \textbf{IDEFICS} \\  \midrule
% epochs                    & 30            & 30             & 30                    & 10                & 10                 & 10                  \\
% batch size                & 8             & 8              & 8                     & 8                 & 8                  & 8                   \\
% early stop                & 5             & 5              & 5                     & -                 & -                 &-                   \\
% learning rate             & 6e-7          & 4e-5           & 4e-5                  & 2e-4              & 5e-5               & 2e-4                \\
% optimizer                 & AdamW         & AdamW          & AdamW                & AdamW             & AdamW              & paged\_adamw\_8bit  \\
% lr schedule               & ExponentialLR & ExponentialLR  & ExponentialLR         & Cosine  & Linear Decay       & Linear Decay  \\ \bottomrule     
% \end{tabular}
% }
% \label{tab:Hyperparameter}
% % \end{table}
% \end{table*}

\section{Annotation Tool}
% 1. 为了提高标注效率，我们开发了一个标注工具。
% 2. 这个工具用来标注，审核，以及测试答题。

\begin{figure}[!t]
\centering
% \vspace{-1.0cm}
\includegraphics[width=\linewidth]{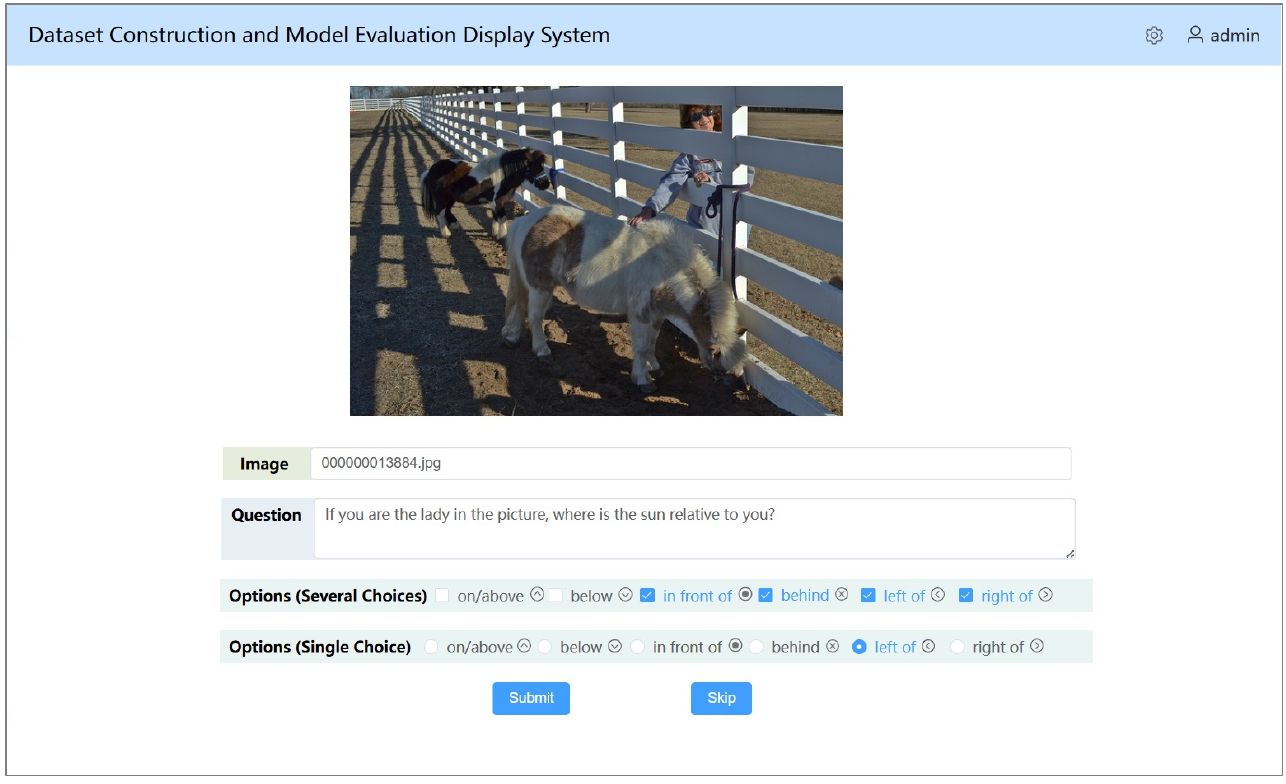}
% \vspace{-0.2cm}
\caption{First-round annotation page in our tool.}
\label{fig:First}
% \vspace{-0.2cm}
\end{figure}

\begin{figure}[!t]
\centering
\vspace{-2.0cm}
\includegraphics[width=\linewidth]{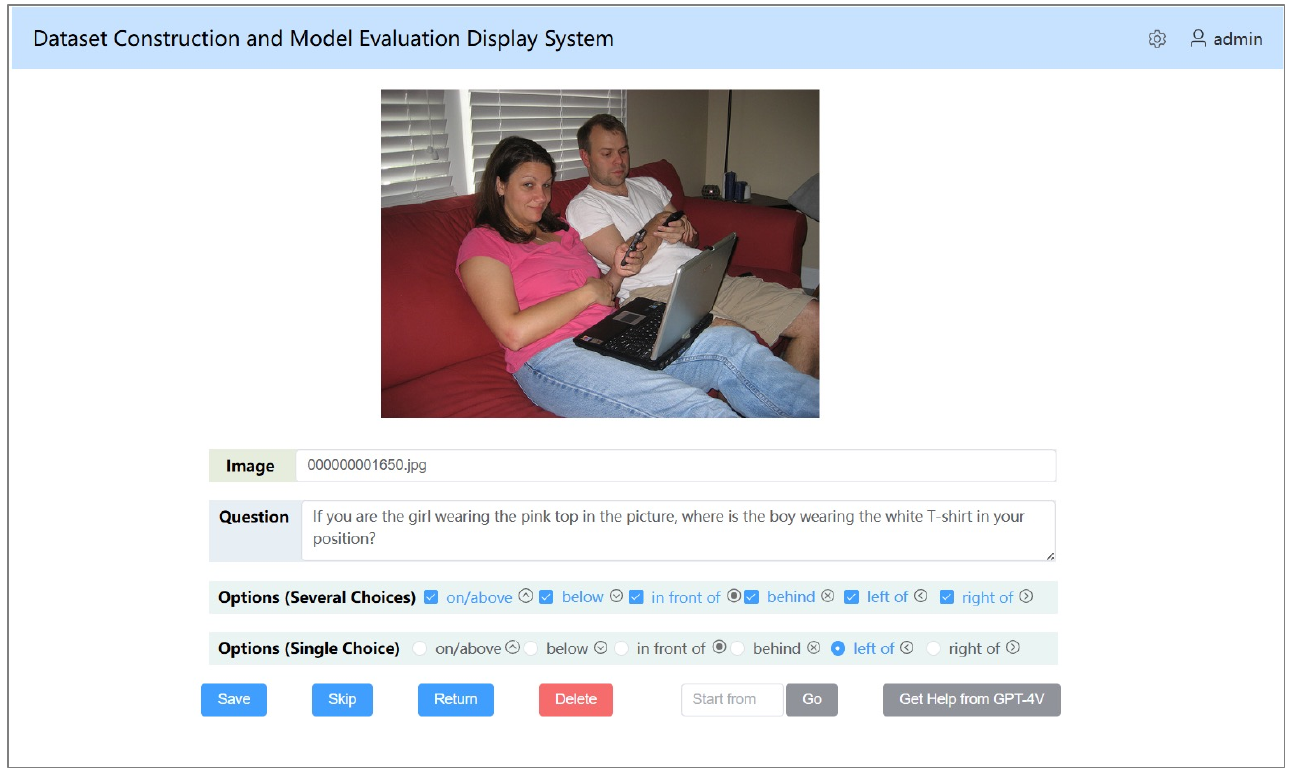}
% \vspace{-0.2cm}
\caption{Second-round checking and third-round review pages in our tool.}
\label{fig:Second}
% \vspace{-3cm}
\end{figure}

\begin{figure}[!t]
\centering
\vspace{-2.0cm}
\includegraphics[width=\linewidth]{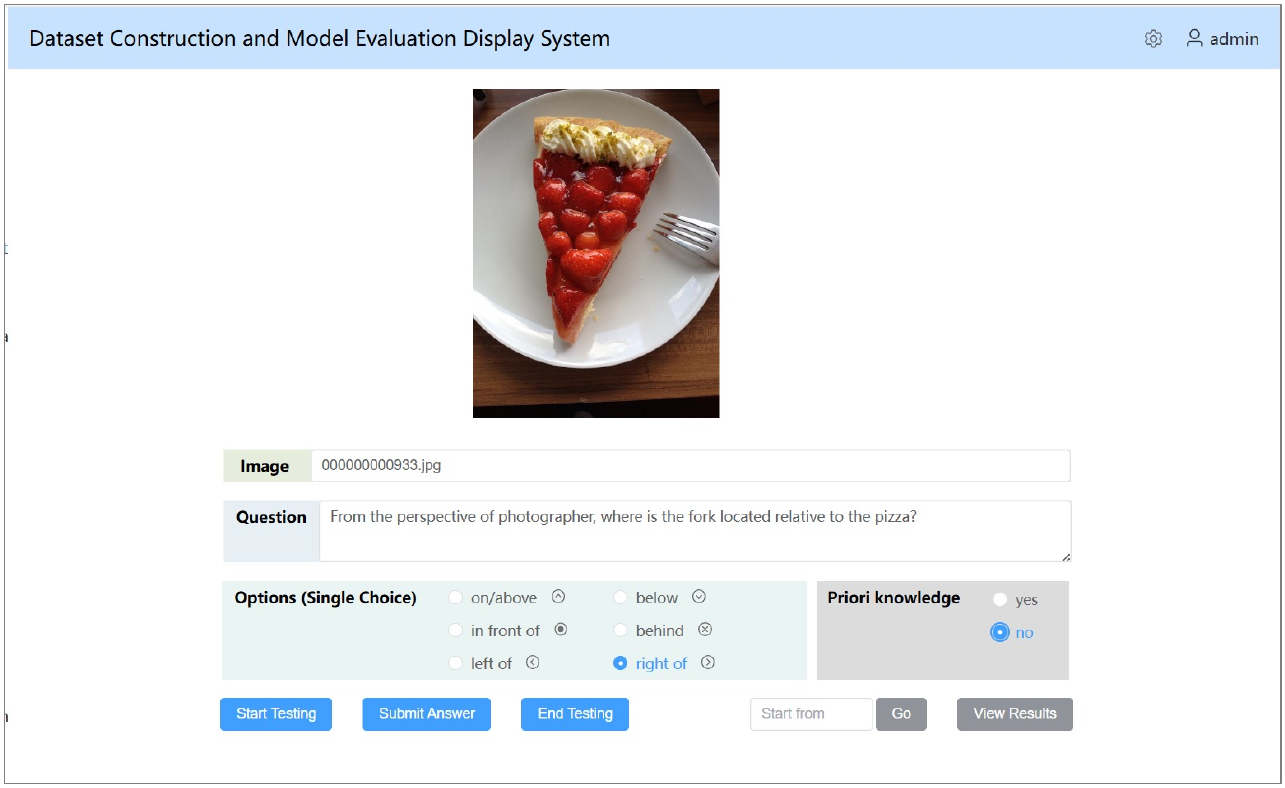}
% \vspace{3cm}
\caption{Human evaluation page in our tool.}
\label{fig:evaluation}
% \vspace{-1.2cm}
\end{figure}

To enhance annotation efficiency, we develop a tool used for annotating (Figure \ref{fig:First}) and checking (Figure \ref{fig:Second}) samples in SpatialMQA, as well as answering (Figure \ref{fig:evaluation}) questions in the test set for evaluators. Each volunteer was compensated at a rate of \$17 per hour.

% \begin{figure*}[!t]
% \centering
% % \vspace{-1.0cm}
% \begin{subfigure}{0.32\linewidth}
%     \centering
%     \includegraphics[width=\linewidth]{11.pdf}
%     % \caption{First-round annotation page}
%     \label{fig:First}
% \end{subfigure}
% \hfill
% \begin{subfigure}{0.32\linewidth}
%     \centering
%     \includegraphics[width=\linewidth]{22.pdf}
%     % \caption{Second-round checking and third-round review page}
%     \label{fig:Second}
% \end{subfigure}
% \hfill
% \begin{subfigure}{0.32\linewidth}
%     \centering
%     \includegraphics[width=\linewidth]{33.pdf}
%     % \caption{Human evaluation page}
%     \label{fig:evaluation}
% \end{subfigure}

% \caption{Interface of the annotation tool. From left to right: First-round annotation page, Second-round checking and third-round review page, Human evaluation page.}
% \label{fig:tool_interface}
% % \vspace{-0.2cm}
% \end{figure*}

\clearpage

\end{document}